\documentclass{article}
\usepackage[T1]{fontenc}
\usepackage{iclr2027_conference,times}

\usepackage{amsmath,amsfonts,bm}

\def\eqref#1{equation~\ref{#1}}
\def\1{\bm{1}}

\DeclareMathAlphabet{\mathsfit}{\encodingdefault}{\sfdefault}{m}{sl}
\SetMathAlphabet{\mathsfit}{bold}{\encodingdefault}{\sfdefault}{bx}{n}

\usepackage{hyperref}
\usepackage{url}
\usepackage{graphicx}
\usepackage{amsmath}
\usepackage{amssymb}
\usepackage{booktabs}
\usepackage{multirow}
\usepackage{float}
\usepackage{xcolor}

\hypersetup{
  pdftitle={StereoGaussians: Feed-Forward 3D Gaussian Splatting from Stereo Images},
  pdfauthor={Boyuan Tian, Huangying Zhan, Zhan Li, Shin-Fang Chng, Hanwen Yang, Zirui Wang, Yi Xu},
  colorlinks=true,
  citecolor=blue,
  linkcolor=red,
  urlcolor=blue,
  pdfborder={0 0 0}
}

\newcommand{\bestscore}[1]{\textbf{#1}}


\def\titlePrefix{StereoGaussians}

\title{\titlePrefix{}: Feed-Forward\\ 3D Gaussian Splatting from Stereo Images}

\author{Boyuan Tian\thanks{Equal contribution.}\quad
Huangying Zhan\textsuperscript{*\,}\thanks{Corresponding author.}\quad
Zhan Li\thanks{Work done while Zhan Li was with Goertek Alpha Labs.}\quad
Shin-Fang Chng \\
\textbf{Hanwen Yang\quad Zirui Wang\quad Yi Xu} \\
Goertek Alpha Labs}

\iclrfinalcopy
\begin{document}

\maketitle
% Override the conference-publication header in the final-copy template.
\lhead{Preprint}

% Keep the opening-page spacing natural instead of stretching it to fill the page.
\begingroup
\raggedbottom
\begin{abstract}
Feed-forward 3D Gaussian Splatting (3DGS) enables reconstruction without
per-scene optimisation, but practical stereo-camera applications require
nearby-view extrapolation beyond the input views. Stereo depth anchors
visible surfaces, yet rendering newly exposed regions also requires
learned appearance and additional scene capacity. We introduce
\titlePrefix{}, which predicts a metric 3DGS representation from a
single calibrated stereo pair. It reuses intermediate representations
from frozen pretrained stereo networks to predict Gaussian attributes,
while calibrated disparity anchors the geometry. A second Gaussian
layer and an expanded image canvas provide capacity for disoccluded
and outside-field-of-view content. For training, we construct
SceneSplat-Stereo from quality-filtered 3DGS teachers, pairing stereo
inputs with nearby target views across 803 training scenes.
Experiments on unseen real and photorealistic stereo benchmarks
demonstrate improvements over strong view-synthesis baselines,
while ablation studies support our main design choices.
\end{abstract}

\vspace{-\baselineskip}
\section{Introduction}

Feed-forward 3D Gaussian Splatting (3DGS) predicts renderable scenes
without per-scene optimisation~\citep{pixelsplat,mvsplat,depthsplat}.
For XR headsets, stereo rigs, and spatial video, a single calibrated
stereo pair must support nearby-view extrapolation, revealing surfaces
hidden from the inputs or extending beyond their field of view.
Unlike interpolation between posed context views, this requires
predicting content outside the observed coverage. Stereo calibration
provides metric geometry for visible surfaces, but accurate matching
alone cannot determine how newly exposed regions render.
Monocular predictors infer geometry and scale from image priors;
calibrated stereo instead supplies a metric reference for learning
a renderable representation.

Our central idea is to transfer pretrained stereo representations
beyond their final disparity output. Stereo networks encode both
correspondence and image context~\citep{foundationstereo,fastfoundationstereo},
providing features from which to learn Gaussian appearance, opacity,
and shape. \titlePrefix{} couples these frozen features with a trainable
predictor: calibrated depth anchors the geometry, while rendering
supervision learns the attributes needed for novel views. A second
Gaussian layer adds capacity behind visible surfaces, and an expanded
prediction canvas adds support outside the input frustum. Together,
these mechanisms connect stereo geometry with learned scene content
for nearby-view extrapolation.
The encoder remains frozen, allowing rendering supervision to train
the Gaussian predictor while preserving the pretrained stereo mapping.

Learning this mapping requires stereo inputs paired with posed target
views of the same static scene. Existing correspondence datasets often
lack this supervision, while synthetic scenes and dynamic captures can
introduce appearance gaps or scene inconsistencies. We address this
training-data gap with SceneSplat-Stereo, rendering calibrated stereo
pairs and nearby targets from quality-filtered real-scene 3DGS teachers.
This combines controlled camera sampling, static-scene consistency,
and reconstructed real-scene appearance.
Target displacements beyond the stereo baseline expose regions
that correspondence supervision alone does not cover.

\begin{samepage}
Our contributions are:
\begin{enumerate}
  \item \textbf{Stereo-to-target training data at scale.}
    SceneSplat-Stereo supplies approximately 324\,K calibrated stereo
    pairs and 3.24\,M target images across 803 training scenes.
    Its teacher-based construction enables consistent rendering
    supervision, supporting stronger cross-dataset generalisation and
    perceptual quality than the evaluated existing stereo training datasets
    on the two novel-view synthesis benchmarks.
  \item \textbf{Stereo representation transfer for Gaussian prediction.}
    \titlePrefix{} maps a single stereo pair to a reusable metric 3DGS
    representation, combining frozen stereo features and depth-anchored
    attributes with two-layer, expanded-canvas prediction for
    disoccluded regions and scene content beyond the reference field of view.
  \item \textbf{State-of-the-art zero-shot view synthesis.}
    \titlePrefix{} achieves state-of-the-art performance among the
    evaluated methods on real and photorealistic stereo benchmarks.
    Ablations establish the contributions of pretrained stereo features,
    two-layer geometry, and expanded prediction coverage to rendering quality.
\end{enumerate}
\end{samepage}

\section{Related Work}
\label{sec:related}

\paragraph{Feed-forward Gaussian prediction.}
pixelSplat~\citep{pixelsplat}, MVSplat~\citep{mvsplat},
MVSGaussian~\citep{mvsgaussian}, and DepthSplat~\citep{depthsplat}
predict Gaussians from posed views. DepthSplat combines multi-view
matching with pretrained monocular depth features to connect depth
estimation and Gaussian prediction. GS-LRM~\citep{gslrm} predicts
Gaussians with transformers, while Long-LRM~\citep{longlrm} scales
to longer image sequences and wider scene coverage.
NoPoSplat~\citep{noposplat}, Splatt3R~\citep{splatt3r}, and
AnySplat~\citep{anysplat} relax camera requirements;
YoNoSplat~\citep{yonosplat} supports both posed and unposed inputs.
ReSplat~\citep{resplat} refines Gaussians through rendering feedback.
Single-image methods such as Splatter Image~\citep{splatterimage},
Flash3D~\citep{flash3d}, and SHARP~\citep{sharp} infer geometry and
appearance from monocular priors, including additional scene capacity
for surfaces not directly visible in the input image.

\paragraph{Transformer-based view synthesis.}
LVSM~\citep{lvsm} synthesises target images using learned latent
tokens or direct input-to-target attention. Efficient-LVSM~\citep{efficientlvsm}
separates input and target processing through decoupled co-refinement
attention. These approaches synthesise views without predicting an
explicit Gaussian scene.

\paragraph{Stereo representations and view synthesis.}
RAFT-Stereo~\citep{raftstereo}, IGEV~\citep{igev}, and
FoundationStereo~\citep{foundationstereo} learn correspondence and
context features for disparity estimation; Fast-FoundationStereo~\citep{fastfoundationstereo}
studies efficient zero-shot stereo. StereoNeRF~\citep{stereonerf}
combines pretrained stereo features with depth-guided plane sweeping
and a stereo depth loss for NeRF rendering from multiple pairs.
Stereo-GS~\citep{stereogs} uses global attention to fuse pairwise stereo
features and separate geometry and appearance heads for pose-free
multi-view Gaussian reconstruction.
WorldMirror~\citep{worldmirror} incorporates camera and depth priors
into joint geometry and Gaussian prediction. In our approach,
calibrated disparity anchors metric geometry, while frozen stereo
features condition Gaussian attributes. Two-layer prediction and an
expanded canvas support nearby-view extrapolation from a single stereo pair.

\paragraph{Stereo and view-synthesis data.}
RealEstate10K~\citep{stereomagnification} and DL3DV~\citep{dl3dv}
provide posed video for view synthesis. Middlebury~\citep{middlebury}
and DrivingStereo~\citep{drivingstereo} benchmark stereo matching,
while Flickr1024~\citep{flickr1024} supports stereo super-resolution.
SceneFlow~\citep{sceneflow} and IRS~\citep{irs} provide synthetic
geometric supervision, but correspondence data need not include posed
target views for rendering supervision. StereoNVS~\citep{stereonerf}
provides stereo-to-target views, and SceneSplat-7K~\citep{li2025scenesplat}
provides real-scene Gaussians. SceneSplat-Stereo builds on these teachers
to jointly control stereo calibration and target-view sampling,
producing appearance-consistent supervision for nearby-view extrapolation.

% Keep the opening two pages dedicated to the abstract, introduction, and related work.
\clearpage
\endgroup
\section{SceneSplat-Stereo Dataset}
\label{sec:dataset}
\label{sec:dataset-requirements}

\begin{figure}[!t]
\setlength{\abovecaptionskip}{0pt}
\setlength{\parskip}{0pt}
\centering
\includegraphics[width=\linewidth]{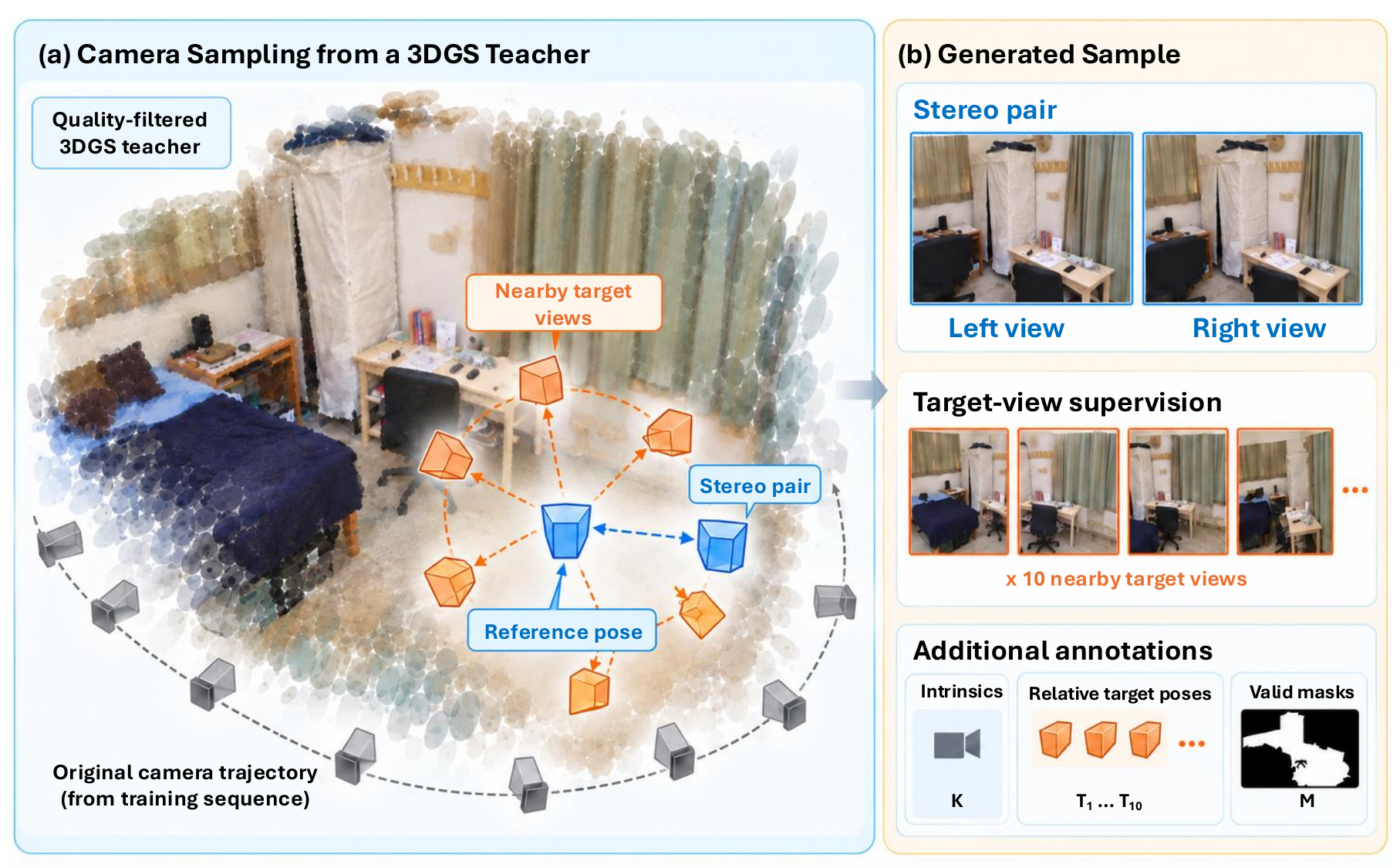}
\caption{\textbf{SceneSplat-Stereo synthetic data generation.}
\textbf{(a) Camera sampling.} A quality-filtered real-scene 3DGS teacher
renders a calibrated stereo pair and ten nearby targets around a
reference pose from the source trajectory.
\textbf{(b) Generated sample.} Stereo inputs and target images are
accompanied by intrinsics, relative target poses, and opacity-derived
valid masks. Camera placement is schematic and illustrates the relative
arrangement of input and target views.}
\label{fig:dataset-generation}
\end{figure}

SceneSplat-Stereo pairs calibrated stereo inputs with nearby target
views rendered from the same static 3DGS teacher. This provides
consistent appearance and known camera poses for learning
stereo-to-3DGS prediction. Figure~\ref{fig:dataset-generation} illustrates
the camera sampling and rendering pipeline, which preserves the
reconstructed teacher's real-scene appearance.
The input pair provides a metric geometric reference, while surrounding
targets supervise rendering as surfaces become visible and image
coverage changes.
Table~\ref{tab:dataset-comparison} compares
its combination of stereo input, target-view supervision, static-scene
consistency, and real-scene appearance with existing datasets used for
stereo matching and novel-view synthesis under different supervision settings.

% Keep the dataset comparison on the next page, after Figure 1.
\suppressfloats[t]
\begin{table}[t]
\centering
\small
\caption{Dataset comparison for feed-forward stereo-to-3DGS
prediction. Scale reports scenes (or sequences) / input images,
counting both stereo views or individual monocular frames;
separately rendered targets are excluded. See Appendix~\ref{app:dataset-comparison}
for details on dataset-specific counting conventions and the
supervision available from each source.}
\label{tab:dataset-comparison}
\setlength{\tabcolsep}{4pt}
\scriptsize
\resizebox{\linewidth}{!}{%
\begin{tabular}{lccccc}
\toprule
Data source & Scenes / images & Stereo input & Target-view sup. & Static scenes & Real appearance \\
\midrule
RealEstate10K~\citep{stereomagnification}
  & 80K / 10M & & $\checkmark$ & $\checkmark$ & $\checkmark$ \\
DL3DV-10K~\citep{dl3dv}
  & 10.5K / 51.2M & & $\checkmark$ & $\checkmark$ & $\checkmark$ \\
Middlebury v3~\citep{middlebury}
  & 30 / 60 & $\checkmark$ & & $\checkmark$ & $\checkmark$ \\
Flickr1024~\citep{flickr1024}
  & 1K / 2K & $\checkmark$ & & $\checkmark$ & $\checkmark$ \\
InStereo2K~\citep{instereo2k}
  & -- / 4.1K & $\checkmark$ & & $\checkmark$ & $\checkmark$ \\
KITTI~\citep{kitti}
  & 789 / 1.6K & $\checkmark$ & & & $\checkmark$ \\
DrivingStereo~\citep{drivingstereo}
  & 42 / 364K & $\checkmark$ & & & $\checkmark$ \\
SceneFlow~\citep{sceneflow}
  & 2.2K / 79.6K & $\checkmark$ & $\checkmark$ & & \\
IRS~\citep{irs}
  & 70 / 207K & $\checkmark$ & & $\checkmark$ & \\
StereoNVS~\citep{stereonerf}
  & 103 / 17.6K & $\checkmark$ & $\checkmark$ & $\checkmark$ & $\checkmark$ \\
\textbf{SceneSplat-Stereo}
  & \textbf{803 / 648K} & \textbf{$\checkmark$} & \textbf{$\checkmark$} & \textbf{$\checkmark$} & \textbf{$\checkmark$} \\
\bottomrule
\end{tabular}
}
\end{table}

\paragraph{Sources and splits.}
\label{sec:dataset-source}
We use the ScanNetGS and ScanNetPPGS-V2 subsets of
SceneSplat-7K~\citep{li2025scenesplat}, selecting them for their scene
scale and capture coverage for nearby rendering. The quality-filtered
pool is split into 803 training and 50 validation scenes, with no scene
overlap. Training contains approximately 324\,K stereo pairs, each with
ten separately rendered targets (3.24\,M target images). For the default
model, StereoNVS and ReplicaGS serve only as test datasets. Source counts and
validation sampling are detailed in Appendix~\ref{app:dataset} and
Table~\ref{tab:dataset-stats}.

\paragraph{Rendering and quality control.}
\label{sec:dataset-construction}
\label{sec:dataset-quality}
For each reference pose, we translate the right camera by baseline $B$
along the local $x$-axis, retaining its orientation and intrinsics to
form a rectified pair. Ten target cameras translate by radius $\rho$
along fixed Fibonacci-sphere directions, with no rotation or
overlap-based rejection. Samples include calibration, relative target
poses, and opacity-derived valid masks, used for ReplicaGS evaluation
but not training.

We retain teachers with released reconstruction PSNR above 30\,dB at
30K optimisation steps, selecting scenes by reconstruction fidelity.
Three quality tiers set $B$ and
$\rho$, with target displacement two to three times the stereo baseline.
Training samples four of the ten targets per pair, exposing the
predictor to varied nearby viewpoints across iterations.
The exact tiers, filtering counts, and rendering protocol are given in
Appendix~\ref{app:dataset}.

\section{\titlePrefix{}}
\label{sec:method}

\subsection{Overview}
\label{sec:method-pipeline}

\begin{figure}[t]
\setlength{\abovecaptionskip}{0pt}
\setlength{\parskip}{0pt}
\centering
% Trim only blank top/bottom PDF margins, keeping panel labels and image scale.
\includegraphics[width=\linewidth,trim=0bp 18bp 0bp 11bp,clip]{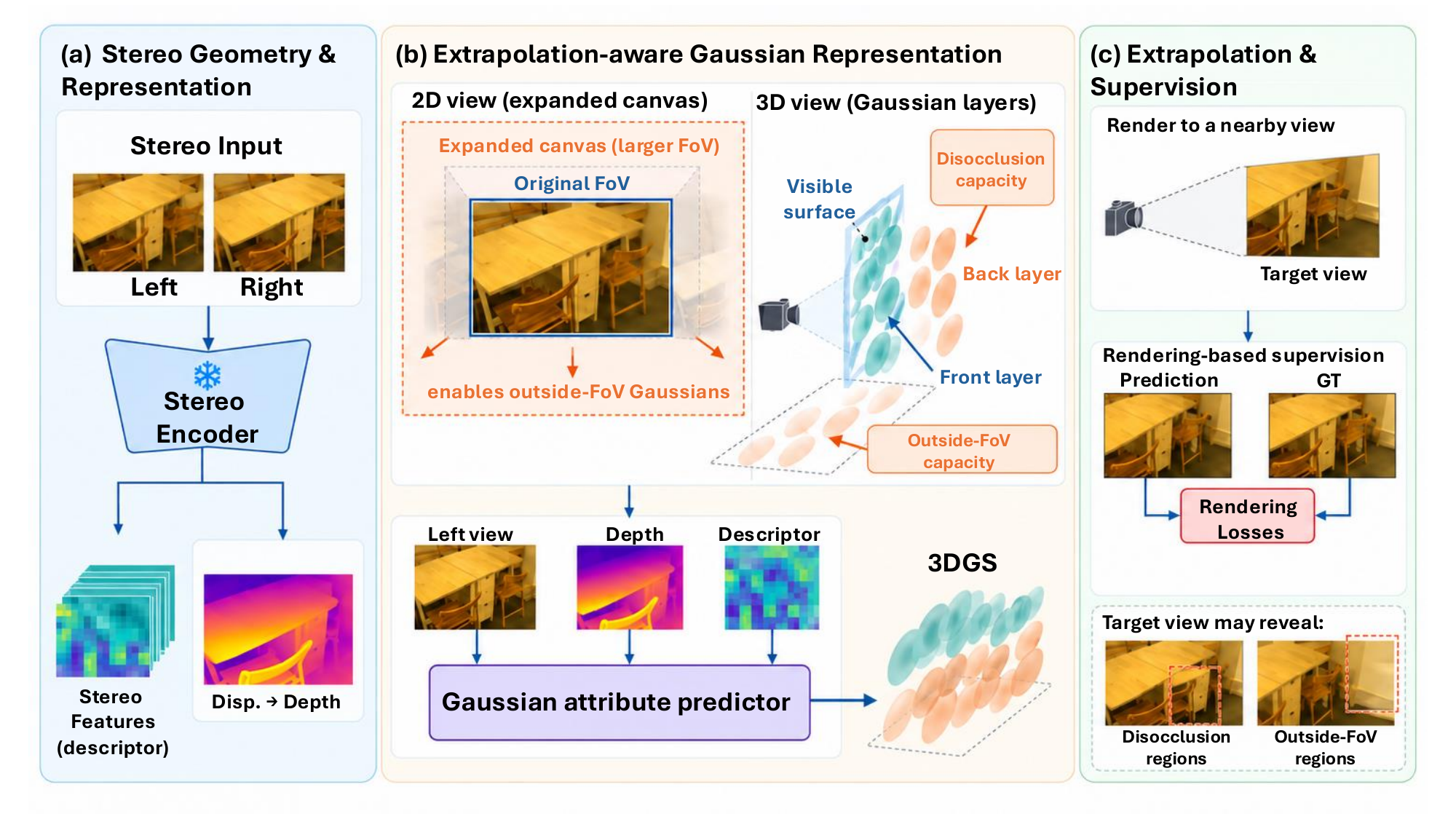}
\caption{\textbf{\titlePrefix{} overview.}
\textbf{(a) Stereo geometry and representation.} Frozen stereo features
and calibrated depth condition Gaussian prediction.
\textbf{(b) Extrapolation-aware representation.} Two layers and an
expanded canvas support disoccluded and outside-FoV content.
\textbf{(c) Extrapolation and supervision.} Target-view rendering trains
the decoder and attribute predictor. Scene and Gaussian illustrations
are schematic.}
\label{fig:framework}
\end{figure}

Given a calibrated rectified stereo pair $(I_L,I_R)$, intrinsics $K$,
and baseline $B$, \titlePrefix{} predicts Gaussians
$\mathcal G=\{(\mathbf x_i,\mathbf s_i,\mathbf q_i,\alpha_i,
\mathbf c_i)\}_{i=1}^{N}$ in the left-camera frame, using the standard
3DGS position, scale, rotation, opacity, and colour
attributes~\citep{3dgs}. A frozen stereo encoder supplies intermediate
features for attribute prediction and disparity for metric geometry.
A trainable predictor constructs two Gaussian layers on an expanded
image canvas, providing capacity for disoccluded and outside-FoV
content. The decoder and predictor learn through differentiable
Gaussian rendering with target-image supervision from SceneSplat-Stereo;
the stereo encoder remains frozen. Figure~\ref{fig:framework} shows the pipeline.

% Keep the framework on the next page rather than stacking it below Table 1.
\suppressfloats[t]

\subsection{Transferring stereo representations to 3DGS}
\label{sec:method-fs}

The encoder exposes a left-reference disparity map $d$ and aligned
multiscale features $\{F^s\}$ through a backbone-specific interface.
A DPT-style decoder $\mathcal D$ concatenates each feature scale with
resized reference RGB and stereo depth $z$ (Eq.~\ref{eq:stereo-depth}),
then projects and fuses the features coarse-to-fine into a dense
full-resolution descriptor for downstream Gaussian attribute prediction:
\begin{equation}
\label{eq:stereo-feature-decoder}
  \Phi=\mathcal D\big(\{[F^s,\downarrow_s I_L,\downarrow_s z]\}_s\big)
  \in\mathbb R^{C\times H\times W}.
\end{equation}
Here $[\cdot,\cdot]$ denotes channel concatenation and $\downarrow_s$
resizes inputs to feature scale $s$. After canvas extension
(\S\ref{sec:method-fov}), a convolutional predictor $\mathcal H$ maps
the extended inputs, denoted by tildes, to attribute residuals for
both layers and a separate back-layer depth offset:
\begin{equation}
\label{eq:gaussian-predictor}
 (\theta_{\mathrm{front}},\theta_{\mathrm{back}},\Delta z)
 =\mathcal H([\tilde I_L,\tilde z,\tilde\Phi]).
\end{equation}
Its shared regressor feeds three output branches. This feature pathway
conditions Gaussian prediction on pretrained stereo representations
beyond the final disparity map. Encoder interfaces and decoder/head
implementation details are given in Appendix~\ref{app:impl}.

\subsection{Geometry-anchored Gaussian prediction}
\label{sec:method-composer}

For reference pixel $p=(u,v)$, calibrated disparity defines metric depth
\begin{equation}
\label{eq:stereo-depth}
  z(p)=\operatorname{clip}\!\left(\frac{f_xB}{d(p)},z_{\min},z_{\max}\right),
\end{equation}
where $f_x$ is the horizontal focal length in pixels. The front layer
retains this processed depth. Each layer $\ell$ predicts a two-channel
output $o_\ell$ giving image-plane offsets
$(\delta_{u,\ell},\delta_{v,\ell})=\sigma(o_\ell)-\tfrac12$,
bounded to half a canvas pixel per axis. Its centre is
\begin{equation}
\label{eq:gaussian-centre}
 \mathbf x_\ell(p)=z_\ell(p)\tilde K^{-1}
 [u+\delta_{u,\ell},v+\delta_{v,\ell},1]^\top,
\end{equation}
where $z_\ell$ and canvas intrinsics $\tilde K$ are specified below.

We adapt SHARP's residual attribute composition~\citep{sharp}:
geometry and reference colour define base attributes, while the
predictor learns bounded multiplicative scale updates, additive
quaternion updates, and opacity updates in logit space. The isotropic
base scale $s_{0,\ell}=\kappa z_\ell/\sqrt{f_xf_y}$ follows the metric
pixel footprint. Our adaptation fixes front-layer stereo depth and
bounds image-plane offsets, whereas SHARP also refines inverse depth
through its composer. A separate branch determines back-layer depth,
with no subsequent depth residual. We also replace RGB composition
with degree-2 spherical-harmonic (SH) residuals around a reference-image
DC base. These constraints provide a structural prior without an
additional supervision loss. Full attribute equations, activations,
and constants are given in Appendix~\ref{app:composer}.

\subsection{Extrapolation-aware prediction}
\label{sec:method-extrapolation}
\label{sec:method-backlayer}
\label{sec:method-fov}

\paragraph{Back layer.}
Both layers have separate attribute outputs at every canvas pixel.
The front layer uses $z_{\mathrm{front}}=\tilde z$; the back layer uses
\begin{equation}
\label{eq:back-layer-depth}
 z_{\mathrm{back}}(p)=\max_{q\in\mathcal N_9(p)}\tilde z(q)
 +\operatorname{softplus}(\Delta z(p)),
\end{equation}
where $\mathcal N_9(p)$ is a $9\times9$ neighbourhood. The local maximum
provides a deeper anchor near depth boundaries, and the learned positive
offset adjusts its depth under rendering supervision. Both the centre
and base footprint use this final depth. We concatenate the two layers;
learned attributes and visibility determine their contributions to a
target view, without an edge mask restricting the back layer to a
predefined subset of image pixels.

\paragraph{Expanded field of view.}
After decoding at the original resolution, we extend $I_L$, $z$, and
$\Phi$ by edge replication to $(H+2P)\times(W+2P)$, where $P$ is the
padding per side. Canvas intrinsics retain the focal lengths and shift
the principal point to $(c_x+P,c_y+P)$, so padded pixels define rays
outside the original image frustum. The predictor operates on the full
canvas, using the extended depth for both layer constructions above.
Target cameras retain their original resolution and intrinsics;
padded Gaussians receive the same rendering supervision through the
target pixels they cover. Canvas implementation details are given in
Appendix~\ref{app:dense-predictor}.

\section{Experiments}
\label{sec:experiments}

% Queue the main comparison early so it can appear above the setup on page 6.
\begin{table}[!t]
\centering
\small
\caption{Zero-shot comparison using the default 16-pixel padding.
Our model is trained on SceneSplat-Stereo. Baselines retain their
respective pretraining. Bold indicates the best score in each column
for the corresponding dataset and evaluation metric.}
\label{tab:main}
\setlength{\tabcolsep}{3pt}
\normalsize
\begin{tabular*}{\linewidth}{@{\extracolsep{\fill}}lccc@{\hspace{10pt}}ccc@{}}
\toprule
& \multicolumn{3}{c}{\textbf{StereoNVS}} & \multicolumn{3}{c}{\textbf{ReplicaGS}} \\
\cmidrule(lr){2-4}\cmidrule(l){5-7}
Method & PSNR $\uparrow$ & SSIM $\uparrow$ & LPIPS $\downarrow$ & PSNR $\uparrow$ & SSIM $\uparrow$ & LPIPS $\downarrow$ \\
\midrule
\multicolumn{7}{c}{\emph{Single-view feed-forward 3DGS}} \\
SHARP-mono & 18.845 & 0.6140 & 0.2604 & 23.388 & 0.7877 & 0.2103 \\
SHARP-stereo & 18.767 & 0.6030 & 0.2788 & 23.363 & 0.7808 & 0.2164 \\
\midrule
\multicolumn{7}{c}{\emph{Sparse-view feed-forward 3DGS}} \\
MVSplat & 17.458 & 0.7248 & 0.2087 & 14.705 & 0.5342 & 0.4152 \\
DepthSplat & 17.977 & 0.7498 & 0.2024 & 16.388 & 0.6161 & 0.3270 \\
\midrule
\multicolumn{7}{c}{\emph{Transformer-based NVS}} \\
Efficient-LVSM & 15.319 & 0.4646 & 0.4752 & 22.780 & 0.7539 & 0.2210 \\
LVSM & 19.796 & 0.6450 & 0.2460 & 25.772 & 0.8409 & 0.1701 \\
\midrule
\multicolumn{7}{c}{\emph{Stereo geometry without learned prediction}} \\
FS-forward-splat & 15.734 & 0.6583 & 0.2891 & 17.017 & 0.7846 & 0.3009 \\
FS-naive-GS & 16.008 & 0.6636 & 0.2802 & 17.918 & 0.8090 & 0.2303 \\
\midrule
\textbf{\titlePrefix{}} & \bestscore{21.261} & \bestscore{0.7549} & \bestscore{0.1653} & \bestscore{30.266} & \bestscore{0.9447} & \bestscore{0.0818} \\
\bottomrule
\end{tabular*}
\end{table}

\subsection{Experimental setup}
\label{sec:setup}

\paragraph{Training and evaluation.}
Our default uses the FoundationStereo encoder and predictor of
\S\ref{sec:method}, with two Gaussian layers, no back-layer edge mask,
and 16-pixel FoV padding. Our architectural ablations train at $256\times256$ for
30K steps with batch size 32, using full-image MSE plus $0.05$ LPIPS
and selecting the checkpoint with highest validation PSNR.
SceneSplat-Stereo pairs are sampled uniformly with four random targets
per pair. Full training and validation details are in
Appendices~\ref{app:impl} and~\ref{app:dataset}, including optimisation,
model selection, and target sampling.

We evaluate zero-shot on StereoNVS (8 real-world scenes, 224 pairs)
and ReplicaGS (8 held-out photorealistic scenes, 800 pairs, 8\,000
targets). Neither test set is used to train or validate the default
SceneSplat-Stereo model. We report
PSNR, SSIM, and LPIPS-VGG; ReplicaGS metrics use opacity-derived valid
masks for teacher-supported regions
(Appendix~\ref{app:eval-details}).

\paragraph{Baselines.}
We compare MVSplat~\citep{mvsplat}, DepthSplat~\citep{depthsplat},
SHARP~\citep{sharp}, LVSM~\citep{lvsm}, and
Efficient-LVSM~\citep{efficientlvsm}. SHARP-mono uses the reference
image; SHARP-stereo concatenates independent left/right predictions
in the reference frame. FS-forward-splat and FS-naive-GS use
FoundationStereo depth with forward splatting and hand-specified
Gaussian attributes, respectively. Table~\ref{tab:main} evaluates released models with
their original pretraining under the zero-shot transfer protocol.
Baseline settings are detailed in
Appendix~\ref{app:baseline-implementation}, including pretrained model
selection and LVSM camera preprocessing.

\subsection{Zero-shot novel-view synthesis}
\label{sec:main-comparison}

\begin{figure}[!t]
\setlength{\abovecaptionskip}{3pt}
\setlength{\parskip}{0pt}
\centering
% Remove blank PDF margins, leaving a small inset around all panel labels.
\includegraphics[width=\linewidth,trim=10bp 42bp 10bp 28bp,clip]{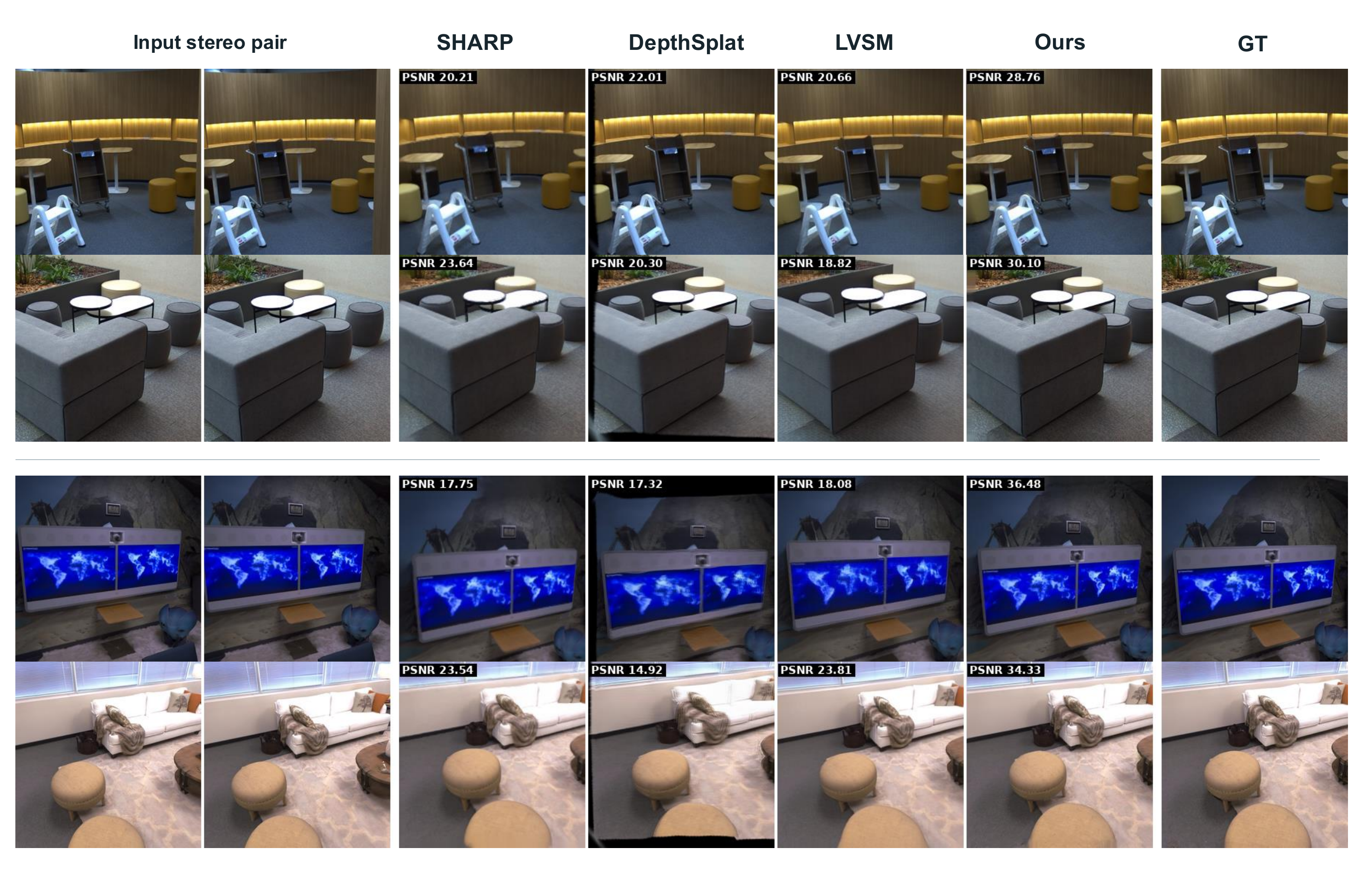}
\caption{\textbf{Qualitative novel-view comparison} on StereoNVS
(top two rows) and ReplicaGS (bottom two rows). Each row shows the input
stereo pair, predictions from representative methods SHARP, DepthSplat,
and LVSM, \titlePrefix{} (Ours), and ground truth (GT). Prediction panels
report PSNR in dB.}
\label{fig:qualitative}
\end{figure}

\titlePrefix{} outperforms all compared baselines on both test sets
across PSNR, SSIM, and LPIPS (Table~\ref{tab:main}). Relative to LVSM,
the strongest released baseline in PSNR, it gains 1.465\,dB
on StereoNVS and 4.494\,dB on ReplicaGS. Both FS baselines
lag behind despite using the same stereo backbone,
indicating that strong geometry alone is insufficient without learned
Gaussian prediction. Qualitative comparisons appear in
Figure~\ref{fig:qualitative} and Appendix~\ref{app:qualitative-comparison}.

\paragraph{Higher-resolution models.}
The same architecture supports $512^2$ and $1024^2$ stereo inputs and
target-view rendering (Appendix~\ref{app:resolution}).

\subsection{Training data and same-data comparison}
\label{sec:training-data}
\begin{table}[!t]
\centering
\small
\caption{Training-data comparison for StereoGaussians (top) and
DepthSplat on SceneSplat-Stereo (bottom). Feature init.: DA-V2 + UniMatch;
FT: RE10K fine-tuning. StereoNVS-to-StereoNVS evaluation is in-domain.
Bold denotes the best out-of-domain result in each column.}
\label{tab:training-data}
\setlength{\tabcolsep}{4pt}
\normalsize
\begin{tabular*}{\linewidth}{@{\extracolsep{\fill}}lccc|ccc@{}}
\toprule
& \multicolumn{3}{c|}{StereoNVS} & \multicolumn{3}{c}{ReplicaGS} \\
Training dataset / model & PSNR $\uparrow$ & SSIM $\uparrow$ & LPIPS $\downarrow$ & PSNR $\uparrow$ & SSIM $\uparrow$ & LPIPS $\downarrow$ \\
\midrule
StereoNVS & 22.707 & 0.7585 & 0.2024 & 23.929 & 0.8914 & 0.1935 \\
IRS & 19.656 & 0.6546 & 0.2412 & 24.614 & 0.8421 & 0.1873 \\
\textbf{SceneSplat-Stereo} & \bestscore{21.261} & \bestscore{0.7549} & \bestscore{0.1653} & \bestscore{30.266} & \bestscore{0.9447} & \bestscore{0.0818} \\
\midrule
DepthSplat (feature init.) & 16.063 & 0.4861 & 0.4018 & 21.764 & 0.7026 & 0.2350 \\
DepthSplat (RE10K FT) & 21.152 & 0.7313 & 0.2057 & 28.996 & 0.9206 & 0.0911 \\
\bottomrule
\end{tabular*}
\end{table}

We train StereoGaussians on three datasets (Table~\ref{tab:training-data}, top).
StereoNVS uses its training split; IRS preparation is in Appendix~\ref{app:training-data}.
On ReplicaGS, SceneSplat-Stereo gains 6.337/5.652\,dB over StereoNVS/IRS
training, with better SSIM and LPIPS. On StereoNVS, it improves LPIPS
relative to in-domain training (0.1653 versus 0.2024), which has higher
PSNR/SSIM, and exceeds IRS on all three metrics. These results support
cross-dataset transfer and perceptual quality.

\paragraph{Same-data method comparison.}
We choose DepthSplat for its shared feed-forward multi-view Gaussian
prediction formulation; LVSM uses transformers to synthesise images
without explicit Gaussians. We train DepthSplat from pretrained depth
features or fine-tune its RE10K model on our data
(Table~\ref{tab:training-data}; Appendix~\ref{app:depthsplat-same-data}).
Fine-tuning gains 3.175/12.608\,dB over the released model on
StereoNVS/ReplicaGS, improving ReplicaGS SSIM/LPIPS but slightly
worsening them on StereoNVS. StereoGaussians leads across all six
metrics, with 1.270\,dB higher ReplicaGS PSNR and 0.0404 lower
StereoNVS LPIPS than fine-tuned DepthSplat. Combined with our ablations,
this supports method benefits beyond the training data alone.

\subsection{Transferring frozen stereo representations}
\label{sec:ablation-channels}
\label{sec:ablations}

Table~\ref{tab:ablations} compares variants with frozen encoders under
the common schedule; Appendix~\ref{app:qualitative-ablation} provides qualitative comparisons on both test datasets.

\paragraph{Feature pathway and RGB-D conditioning.}
Removing the DPT descriptor retains RGB and stereo geometry but reduces
PSNR by 0.509/1.982\,dB on StereoNVS/ReplicaGS, with worse SSIM and
LPIPS (block~A). Figure~\ref{fig:composer-diagnostic} shows corresponding
tabletop distortions and foreground-boundary artifacts.
RGB-D conditioning improves LPIPS on both benchmarks and ReplicaGS
PSNR by 0.313\,dB, with little change in StereoNVS PSNR. This ablation
removes RGB-D from the decoder and predictor, retaining stereo features
and metric depth anchors.

\paragraph{Different frozen encoders.}
\label{sec:ablation-stereo-backbone}
Our framework also transfers pretrained representations from
Fast-FoundationStereo~\citep{fastfoundationstereo}. With an adapted
feature interface and a separately trained predictor, this variant
reaches 20.876/29.404\,dB on StereoNVS/ReplicaGS (block~B), exceeding
the released external baselines in PSNR and LPIPS on both sets.
Implementation details for both encoders are in Appendix~\ref{app:impl}.

\begin{table}[!t]
\centering
\small
\caption{Ablations of input conditioning, frozen stereo encoders,
Gaussian prediction, and FoV expansion under a shared training and
zero-shot evaluation protocol. Each block varies the default in
\S\ref{sec:setup}; variant definitions are given in
\S\ref{sec:ablations}--\ref{sec:ablation-arch}.}
\label{tab:ablations}
\label{tab:input-channels}
\label{tab:stereo-backbone}
\label{tab:arch-ablation}
\label{tab:fov-sweep}
\setlength{\tabcolsep}{3pt}
\resizebox{\linewidth}{!}{%
\begin{tabular}{lccc|ccc}
\toprule
& \multicolumn{3}{c|}{StereoNVS} & \multicolumn{3}{c}{ReplicaGS} \\
Variant & PSNR $\uparrow$ & SSIM $\uparrow$ & LPIPS $\downarrow$ & PSNR $\uparrow$ & SSIM $\uparrow$ & LPIPS $\downarrow$ \\
\midrule
\textbf{Shared default} & 21.261 & 0.7549 & 0.1653 & \bestscore{30.266} & 0.9447 & 0.0818 \\
\midrule
\multicolumn{7}{l}{\textbf{(A) Input conditioning}} \\
Without descriptor & 20.752 & 0.7417 & 0.1871 & 28.284 & 0.9362 & 0.1045 \\
Without RGB-D conditioning & 21.269 & 0.7594 & 0.1717 & 29.953 & 0.9391 & 0.0915 \\
\midrule
\multicolumn{7}{l}{\textbf{(B) Frozen stereo encoder}} \\
Fast-FoundationStereo & 20.876 & 0.7409 & 0.1737 & 29.404 & 0.9368 & 0.0876 \\
\midrule
\multicolumn{7}{l}{\textbf{(C) Gaussian prediction design}} \\
Without back layer & 21.081 & 0.7506 & 0.1698 & 26.440 & 0.9255 & 0.0988 \\
Hard edge mask & 21.351 & 0.7573 & 0.1643 & 27.421 & 0.9314 & 0.0951 \\
Soft edge mask & 21.125 & 0.7527 & 0.1662 & 30.256 & \bestscore{0.9452} & \bestscore{0.0816} \\
Without composer & 21.875 & 0.7510 & 0.1887 & 30.029 & 0.9445 & 0.0845 \\
\midrule
\multicolumn{7}{l}{\textbf{(D) FoV padding per side}} \\
without padding & 18.093 & 0.7238 & 0.1956 & 21.534 & 0.8884 & 0.1393 \\
32 pixels & 22.400 & 0.7600 & 0.1639 & 30.188 & 0.9445 & 0.0823 \\
64 pixels & \bestscore{23.189} & \bestscore{0.7689} & \bestscore{0.1612} & 30.165 & 0.9443 & 0.0823 \\
\bottomrule
\end{tabular}%
}
\end{table}

\subsection{Gaussian prediction design}
\label{sec:ablation-arch}
\label{sec:ablation-backlayer}
\label{sec:ablation-composer}

\begin{figure}[!t]
\setlength{\abovecaptionskip}{3pt}
\setlength{\parskip}{0pt}
\centering
% Remove only the blank outer margins; preserve annotations and panel labels.
\includegraphics[width=\linewidth,trim=14bp 5bp 14bp 4bp,clip]{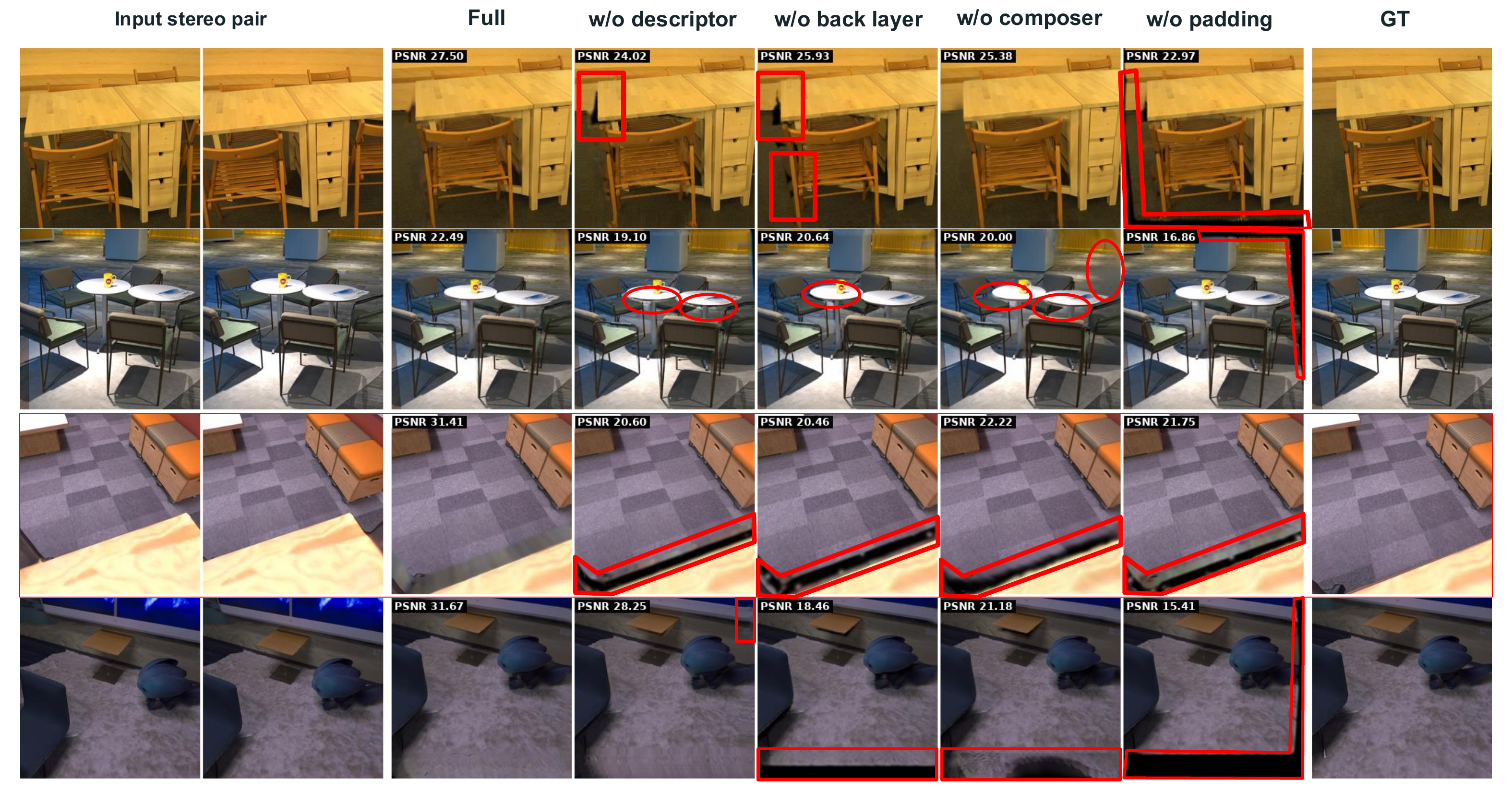}
\caption{\textbf{Qualitative ablations.} Each row shows the input stereo
pair, the full model, variants without the feature descriptor, back
layer, composer, or FoV padding, and ground truth (GT). Red outlines
highlight missing content and distortions near occlusion boundaries
and in newly exposed image-boundary regions; corresponding full-model
and GT regions provide visual references. Prediction panels report
PSNR in dB.}
\label{fig:composer-diagnostic}
\end{figure}

\paragraph{Second-layer capacity.}
Removing the back layer retains stereo-anchored front Gaussians but reduces
PSNR by 0.180/3.826\,dB on StereoNVS/ReplicaGS
(Table~\ref{tab:ablations}, block~C). The desk/chair regions in
Figure~\ref{fig:composer-diagnostic} show missing content and disrupted
occlusion boundaries; the full model better reconstructs these exposed regions.

\paragraph{Hard and soft edge masks.}
We test back-layer opacity masks: hard masks zero opacity outside
depth-edge regions, while soft masks attenuate it
(Appendix~\ref{app:edge-masks}). Qualitatively, both promote clearer
front/back separation and back-layer specialisation in occluded
content. Rendering gains vary across datasets
(Table~\ref{tab:ablations}, block~C), so we use the simpler ungated
design as the default.

\paragraph{Attribute composition.}
The attribute priors in \S\ref{sec:method-composer} improve LPIPS on
both datasets and all three ReplicaGS metrics. Composition lowers
StereoNVS LPIPS from 0.1887 to 0.1653 and improves SSIM, while the
composer-free variant has 0.614\,dB higher PSNR. ReplicaGS PSNR rises
by 0.237\,dB. We retain composition for its consistent perceptual gains
on both benchmarks.

\paragraph{Field-of-view expansion.}
\label{sec:ablation-fov}
At fixed $256^2$ resolution, removing 16-pixel padding lowers PSNR by
3.168/8.732\,dB on StereoNVS/ReplicaGS and worsens SSIM and LPIPS
(block~D). Using 64 pixels instead improves StereoNVS PSNR/LPIPS by
1.928\,dB/0.0034 but lowers ReplicaGS PSNR by 0.101\,dB.
Our 16-pixel default uses a smaller canvas and fewer Gaussians.
Figure~\ref{fig:composer-diagnostic} shows empty target-border bands
without padding, which the full model fills more completely.

\section{Conclusion and Discussion}
\label{sec:conclusion}

\titlePrefix{} transfers frozen stereo features into metric 3DGS
prediction from a single calibrated pair. Two-layer prediction and an
expanded canvas support nearby-view extrapolation, while
SceneSplat-Stereo supplies consistent target-view supervision from
static 3DGS teachers. The model achieves state-of-the-art performance
among evaluated methods on StereoNVS and ReplicaGS. Experiments
demonstrate the effectiveness of SceneSplat-Stereo for cross-dataset
generalisation, while ablations establish the benefits of stereo
feature transfer and prediction capacity for newly exposed regions.

DepthSplat also benefits from SceneSplat-Stereo fine-tuning, while
StereoGaussians retains better test scores. Together with transfer from
both frozen encoders, this supports a practical route from calibrated
stereo to reusable Gaussian scenes.

\paragraph{Limitations and directions.}
Fixed front-layer depth limits stereo-error correction, and teacher
artifacts can persist despite reconstruction-quality filtering.
Our evaluation focuses on nearby views. Future work includes geometry
refinement, supervision from real target photographs, and extending
evaluation to larger viewpoint changes and region-specific measures
of rendering quality in disoccluded and newly exposed outside-FoV regions.

\section*{AI Use Statement}

We used generative AI tools to assist with manuscript preparation,
including improving grammar, wording, clarity, and organisation.
We also used these tools to create selected illustrative components
in Figures~\ref{fig:dataset-generation} and~\ref{fig:framework}, which
serve explanatory purposes and do not represent experimental outputs.
In addition, AI tools assisted with experimental code development and
optimisation, as well as the execution of experiments. The reported
numerical results were obtained by running the experimental code.
We reviewed the AI-assisted code changes and verified the reported
experimental results against the corresponding experimental outputs.
We take full responsibility for the final manuscript, experimental
methodology, implementation, and reported results.

\section*{Reproducibility Statement}

Section~\ref{sec:dataset} and Appendix~\ref{app:dataset} describe the
construction of SceneSplat-Stereo, including source-scene selection,
data splits, camera sampling, and rendering settings.
Section~\ref{sec:method} and Appendix~\ref{app:impl} specify the model
architecture, frozen stereo features, Gaussian parameterisation,
and training hyperparameters. Section~\ref{sec:setup} and
Appendices~\ref{app:eval-details}--\ref{app:baseline-implementation}
describe the evaluation datasets, metrics, and baseline configurations.
Appendices~\ref{app:resolution}--\ref{app:training-data} provide
additional details on higher-resolution experiments and IRS data
preparation.

\bibliography{references}
\bibliographystyle{iclr2027_conference}

\clearpage
\appendix

\section{Implementation details}
\label{app:impl}

\paragraph{Optimisation.}
We optimise the trainable decoder and predictor using AdamW with a
learning rate of $4\times10^{-4}$, weight decay $0.01$, and momentum
coefficients $(\beta_1,\beta_2)=(0.9,0.999)$. The OneCycleLR schedule
uses 2\,700 warm-up steps (9\% of training), with initial and final
learning-rate division factors of 25 and 100, respectively.

\paragraph{Training and model selection.}
Training lasts 30\,000 steps, with validation every 3\,000 steps.
All test results use the model selected by the highest validation PSNR.

\paragraph{Rendering objective.}
Given a target image $I_T$ and its rendered prediction $\hat I_T$,
we minimise $\mathcal{L}=\mathrm{MSE}(I_T,\hat I_T)+
0.05\,\mathrm{LPIPS}_{\mathrm{VGG}}(I_T,\hat I_T)$.
Supervision is applied to target views.

\paragraph{Stereo feature extraction.}
The frozen FoundationStereo encoder supplies disparity and image,
context, and recurrent features aligned with the left reference view.
These outputs provide the geometric anchors and decoder inputs used
for Gaussian prediction. Input RGB values in $[0,1]$ are rescaled to
$[0,255]$ and undergo the encoder's ImageNet normalisation.

\paragraph{Fast-FoundationStereo variant.}
We use a frozen pretrained Fast-FoundationStereo model\footnote{Pretrained
model identifier: \texttt{23-36-37}.} with a single GRU layer, eight
recurrent updates, and PyTorch cost-volume construction. Its decoder
uses left-reference image features and RGB-D at each scale; context
and recurrent features are not included. This feature selection differs
from the FoundationStereo variant described above. Both variants in
Table~\ref{tab:ablations}, block~B, use the two-layer geometry and
attribute composition of \S\ref{sec:method}, with unmasked back-layer
opacity and 16-pixel canvas padding. Each predictor is trained
separately at $256\times256$ with batch size 32 for 30\,000 steps.

\subsection{Dense predictor and canvas}
\label{app:dense-predictor}

At each feature scale, the decoder concatenates the available encoder
features with resized RGB and depth, projecting them to 128 channels
with a $1\times1$ convolution, group normalisation, and GELU. RGB is
resized bilinearly and depth by nearest-neighbour interpolation.
In Eq.~\ref{eq:stereo-feature-decoder}, $s$ is the spatial downsampling
factor, so inputs at scale $s=4$ have size $H/4\times W/4$.
Residual convolution units and coarse-to-fine fusion blocks yield a
128-channel descriptor at the original image resolution.

After edge-replication padding, a regressor with two $3\times3$
convolutions and an intermediate GELU maps
$[\tilde I_L,\tilde z,\tilde\Phi]$ to a 128-channel tensor $h$.
Each output branch processes $[h,\tilde I_L,\tilde\Phi]$ using a
$3\times3$ convolution, group normalisation, GELU, and a final
$3\times3$ convolution. The front and back branches each output
37 channels: opacity (1), image-plane offset (2), scale (3), quaternion
(4), and degree-2 SH coefficients (27). A separate branch outputs the
scalar back-layer depth offset.

\paragraph{Canvas geometry.}
Only the regressor and output branches operate on the expanded canvas;
the dense decoder operates at the original resolution. Edge replication
extends the reference RGB, processed depth, and descriptor, so padded
inputs inherit the nearest border pixel's values. In pixel coordinates,
the focal lengths are unchanged and the principal point shifts as
\begin{equation}
\label{eq:canvas-intrinsics}
 \tilde c_x=c_x+P,\qquad \tilde c_y=c_y+P.
\end{equation}
The front layer uses the extended depth directly, and the back layer
applies Eq.~\ref{eq:back-layer-depth} to that same extended map.
Each layer's centre follows its sub-pixel-shifted canvas ray in
Eq.~\ref{eq:gaussian-centre}.
The predictor uses replicate-padded convolutions and the RGB, depth,
and descriptor inputs specified in Eq.~\ref{eq:gaussian-predictor}.
The same attribute constraints and rendering objective apply across
the original and extended regions. In normalised intrinsics, canvas
expansion gives
$\tilde f_x^n=f_x^nW/(W+2P)$ and
$\tilde c_x^n=(c_x^nW+P)/(W+2P)$, with analogous vertical expressions.
The target cameras retain their own intrinsics and original resolution.

\subsection{Attribute composition}
\label{app:composer}

The following specifies our adaptation of SHARP's residual attribute
composition~\citep{sharp}, including the geometry constraints, SH
parameterisation, and numerical settings used in our model.

For each layer $\ell$, the attribute branch outputs image-plane offsets
$o_\ell$ and residuals
$(\boldsymbol\delta_{s,\ell},\boldsymbol\delta_{q,\ell},
\delta_{\alpha,\ell},\boldsymbol\delta_{c,\ell})$.
The centre is defined in Eq.~\ref{eq:gaussian-centre}; the remaining
attributes are composed as
\begin{subequations}
\label{eq:attribute-composition}
\begin{align}
 \mathbf s_\ell(p)&=\mathrm{clip}\big(s_{0,\ell}(p)
    \phi(\boldsymbol\delta_{s,\ell}),s_{\min},s_{\max}\big),
    \label{eq:gaussian-scale}\\
 \mathbf q_\ell(p)&=\mathrm{normalise}(\mathbf q_{\mathrm{id}}
    +r\boldsymbol\delta_{q,\ell}),\label{eq:gaussian-rotation}\\
 \alpha_\ell(p)&=\sigma(\operatorname{logit}(\alpha_0)
    +\delta_{\alpha,\ell}),\label{eq:gaussian-opacity}\\
 \mathbf c_{0,\ell}(p)&=\mathrm{RGB2SH}(\tilde I_L(p))
    +\boldsymbol\delta_{c,0,\ell}.\label{eq:gaussian-colour}
\end{align}
\end{subequations}
Here $s_{0,\ell}=\kappa z_\ell/\sqrt{f_xf_y}$ is an isotropic base
scale. The bounded sigmoid multiplier $\phi$ acts independently on each
axis, allowing anisotropic scales; $\mathbf q_{\mathrm{id}}$ is the
identity quaternion, $r$ scales rotation residuals, and $\alpha_0$ is
the opacity base. The final equation specifies the DC term of degree-2
SH appearance. At zero attribute residual, each layer uses its
depth-defined centre, base footprint, and reference-image colour.

Stereo z-depth is clipped to $[0.1,10]$\,m before canvas extension.
For each layer, the image-plane displacement is
$\sigma(o)-\tfrac12$, bounded to half a canvas pixel on each axis.
The front layer uses the processed stereo depth directly. Back-layer
depth follows Eq.~\ref{eq:back-layer-depth}. Its depth branch is
zero-initialised, so the initial increment above the pooled depth is
$\ln 2\approx0.69$\,m.

\paragraph{Scale and rotation.}
The isotropic base scale is $s_0=0.5z_\ell/\sqrt{f_xf_y}$, with focal
lengths in pixels. The three predicted scales are
$\mathbf s=\mathrm{clip}(s_0\phi(\boldsymbol\delta_s),10^{-10},3)$
in metres, where
\[
 \phi(t)=4.8\,\sigma(1.5t+\ln0.2)+0.2.
\]
This multiplier has $\phi(0)=1$ and $\phi'(0)=1$.
Rotation uses $r=0.1$ and $\mathbf q_{\mathrm{id}}=(0,0,0,1)$;
the covariance is $R(\mathbf q)\operatorname{diag}(\mathbf s)^2
R(\mathbf q)^\top$ in the reference-camera frame.

\paragraph{Opacity and colour.}
The opacity base is $\alpha_0=0.99$. For degree-2 SH, the DC base is
$\mathrm{RGB2SH}(c)=(c-0.5)/C_0$, with $C_0\approx0.2821$.
Higher-order SH residuals at degree $j$ are scaled by
$0.1\cdot0.25^j$. Both layers use the same reference-image colour
base and separate attribute outputs. Within the attribute branches, only scale and SH output channels
are zero-initialised; opacity, image-plane offset, and quaternion
outputs start small but nonzero. The initial Gaussian attributes
therefore include small deviations from their zero-residual values.

\paragraph{Scale-residual clamps.}
Before composition, raw scale residuals are upper-bounded as
$\boldsymbol\delta_s\leftarrow
\min(\boldsymbol\delta_s,c)$, where
$c=4+\ln(\exp(s_{\mathrm{cap}})-1)$.
For front-layer Gaussians, $s_{\mathrm{cap}}=0.3$ within a dilated
disparity-edge band (relative-gradient threshold 0.05, dilation 8 pixels).
For all back-layer Gaussians, $s_{\mathrm{cap}}=0.5$.
These thresholds bound the scale multipliers to approximately 4.73
and 4.89, respectively, close to the composer's bound of 5. They
constrain dimensionless multipliers of the depth-dependent base scale.
The default model uses unmasked back-layer opacity.

\paragraph{Edge-mask variants.}
\label{app:edge-masks}
For the design exploration in \S\ref{sec:ablation-composer}, both
masking variants retain two Gaussian layers and gate only back-layer
opacity. For a binary disparity-edge-region mask $M$, hard masking uses
$\alpha'_{\mathrm{back}}=M\alpha_{\mathrm{back}}$; soft masking uses
$\alpha'_{\mathrm{back}}=[\beta+(1-\beta)M]\alpha_{\mathrm{back}}$
with $\beta=0.3$, leaving opacity unchanged inside the edge region and
scaling it to 30\% of its original value outside. These opacity
gates differ from the scale-residual clamps above and are not used
in the default model.

\section{Dataset generation details}
\label{app:dataset}

\paragraph{Dataset comparison conventions.}
\label{app:dataset-comparison}
Table~\ref{tab:dataset-comparison} uses dataset-specific counting scopes:
KITTI combines the 2012 and 2015 train/test sets; SceneFlow counts
training scenes only; and IRS counts scene instances. StereoNVS
combines real and synthetic subsets, with approximate counts.
SceneSplat-Stereo reports its training split, and -- denotes an
unavailable scene count. The data properties also require several
qualifications. Flickr1024 stereo pairs are uncalibrated. IRS has no
released camera poses for target-view supervision; our pose-recovery
procedure is described in Appendix~\ref{app:training-data}.
StereoNVS's real-appearance checkmark applies only to its real subset.

\begin{table}[!t]
\centering
\small
\caption{Dataset composition. Slash-separated counts denote train / val;
training and validation are scene-disjoint. ReplicaGS is test-only.
Pair and target counts describe the stored training pool or the selected
validation/test views; training draws four of the ten stored targets per
pair. Counts with K/M are rounded. Teacher PSNR averages over the
quality-filtered source pool (train and validation together), or the
eight ReplicaGS test scenes.}
\label{tab:dataset-stats}
\setlength{\tabcolsep}{4pt}
\resizebox{\linewidth}{!}{%
\begin{tabular}{llrrrr}
\toprule
Subset & Split & Scenes & Stereo pairs & Target views & Teacher PSNR \\
\midrule
ScanNetGS & Train / Val & 491 / 30 & 175K / 150 & 1.75M / 600 & 31.69 \\
ScanNetPPGS-V2 & Train / Val & 312 / 20 & 149K / 100 & 1.49M / 400 & 33.04 \\
\textbf{Total} & Train / Val & \textbf{803 / 50} & \textbf{324K / 250} & \textbf{3.24M / 1K} & -- \\
\midrule
ReplicaGS & Test & 8 & 800 & 8K & 41.21 \\
\bottomrule
\end{tabular}
}
\end{table}

\paragraph{Source selection.}
We select ScanNetGS and ScanNetPPGS-V2 from SceneSplat-7K for their
scene scale and coverage for nearby rendering. This selection targets
scenes with sufficient capture coverage to support nearby novel views.
Filtering by released teacher PSNR
strictly above 30\,dB retains 521 of 1\,613 ScanNetGS scenes and
332 of 956 ScanNetPPGS-V2 scenes. Table~\ref{tab:dataset-stats}
summarises the resulting training, validation, and test counts.

\paragraph{Scene splits and counts.}
The quality-filtered pool contains 853 scenes. With seed 42, we sample
30 of the 521 ScanNetGS scenes and 20 of the 332 ScanNetPPGS-V2 scenes
uniformly without replacement for validation, assigning the remainder
to training. ScanNetGS contributes 174\,621 training stereo pairs and
1\,746\,210 targets; ScanNetPPGS-V2 contributes 149\,196 pairs and
1\,491\,960 targets. Training samples pairs uniformly from the combined
323\,817-pair pool. Validation
uses five evenly spaced pairs per held-out scene and the first four
stored targets per pair, for 250 pairs and 1\,000 targets in total.

\paragraph{Reference and target cameras.}
Reference poses follow the complete supplied camera trajectories in
their original order. For ScanNetGS and ReplicaGS, these coincide with
the teacher training poses. For ScanNetPPGS-V2, 80.6\% are teacher
training poses and 19.4\% correspond to frames excluded from teacher
fitting by blur filtering; none are teacher evaluation poses. All
reference images are rendered from the teachers at these poses.
Let $T_L$ be the reference
camera-to-world transform. The right camera is
$T_R=T_L\operatorname{Trans}(B,0,0)$, and target $j$ is
$T_j=T_L\operatorname{Trans}(\rho\mathbf u_j)$, where the ten unit
directions $\mathbf u_j$ are fixed Fibonacci-sphere directions.
All cameras share the reference orientation. Directions are shared
across frames and scenes; only $B$ and $\rho$ vary with teacher quality.
Target generation does not reject views based on overlap.

\paragraph{Quality tiers.}
Teacher quality is the released per-scene reconstruction PSNR at 30K
optimisation steps. ScanNet evaluates the teacher on its training views.
ScanNet++ evaluates on DSLR test frames that are also included in
teacher fitting through its combined train-and-test split.
The threshold therefore measures reconstruction fidelity,
not held-out-view generalisation. These within-scene teacher protocols
are distinct from our scene-disjoint split: all generated views of a
scene belong to the same student training, validation, or test set.
We retain scenes strictly above 30\,dB and apply the
baseline and target-radius tiers in Table~\ref{tab:quality-tiers}.
\begin{table}[!t]
\centering
\small
\caption{Teacher-quality tiers, stereo baselines, target radii, and scene
counts. Scene counts are ordered as ScanNetGS / ScanNetPPGS-V2.}
\label{tab:quality-tiers}
\begin{tabular}{lcccc}
\toprule
Teacher PSNR & $B$ (m) & $\rho$ (m) & Train scenes & Val scenes \\
\midrule
$(30,35)$\,dB & 0.05 & 0.10 & 473 / 257 & 28 / 14 \\
$[35,40)$\,dB & 0.10 & 0.20 & 18 / 48 & 2 / 6 \\
$[40,\infty)$\,dB & 0.10 & 0.30 & 0 / 7 & 0 / 0 \\
\bottomrule
\end{tabular}
\end{table}
The smallest tier contains 730 of the 803 training scenes (approximately
91\%). All eight ReplicaGS test scenes use $B=0.10$\,m and
$\rho=0.20$ or $0.30$\,m according to the same rule.

\paragraph{Rendering and valid masks.}
Images are rendered using antialiased gsplat at the teachers' native
resolutions: $616\times456$ for ScanNetGS, $1752\times1168$ for
ScanNetPPGS-V2, and $1200\times680$ for ReplicaGS, then stored as JPEG
with quality 92. Camera poses are expressed in a common OpenCV
convention, including the corresponding axis conversion for
ScanNet++ sources. Each target includes a valid mask,
defined by rendered alpha $\geq0.5$. The training objective is
full-image MSE plus $0.05$ LPIPS and does not use this mask. Four
targets are selected randomly from the ten stored views per training
sample. ReplicaGS evaluation uses all ten and applies valid masks.

\section{Evaluation-set details}
\label{app:eval-details}

\paragraph{StereoNVS.} We use the 8 paper-holdout real-world scenes
from StereoNVS-Real, giving 224 calibrated stereo-to-target evaluation
pairs with COLMAP-estimated camera poses. For the default model, this
benchmark tests zero-shot transfer to captured stereo imagery.

\paragraph{ReplicaGS.} ReplicaGS is used only for evaluation. It
contains 8 held-out photorealistic Replica scenes rendered from
optimised 3DGS teachers using known rendering camera parameters. Each scene provides
a 2\,000-frame trajectory; we select 100 evenly spaced reference poses
per scene, yielding 800 stereo pairs, and render ten nearby target
views per pair, yielding $8\times100\times10=8\,000$ target views.
The mean teacher PSNR across the eight scenes is 41.21\,dB.
PSNR, SSIM, and LPIPS are evaluated within a valid mask marking
target regions supported by the teacher rendering (rendered alpha
$\geq0.5$).

\paragraph{Camera accuracy and image alignment.}
ReplicaGS uses known rendering intrinsics and extrinsics, while
StereoNVS uses captured images with estimated camera poses. Residual
calibration or pose errors can affect image alignment and pixelwise
metrics on StereoNVS. We compare methods within each benchmark under
its evaluation protocol, accounting for their distinct image and
camera sources.

\section{Baseline implementation details}
\label{app:baseline-implementation}

\paragraph{Pretrained model selection.}
Both SHARP-mono and SHARP-stereo use SHARP's official pretrained
checkpoint. For MVSplat, DepthSplat, LVSM, and Efficient-LVSM, we use
their respective $256\times256$ versions.

\paragraph{LVSM camera preprocessing.}
Following LVSM's camera normalisation, we recenter and rotation-align
the poses and rescale translations by
$1/(1.35\,\|t\|_{\mathrm{max}})$.
Evaluation uses the designated target views for each input pair.

\paragraph{DepthSplat trained on SceneSplat-Stereo.}
\label{app:depthsplat-same-data}
Table~\ref{tab:training-data} includes two training routes. The feature-init.
variant initialises the monocular branch with Depth Anything V2 and the
multi-view branch with UniMatch, following DepthSplat's feature
initialisation recipe~\citep{depthsplat}, then trains on SceneSplat-Stereo
without RE10K view-synthesis training. The RE10K FT variant instead
fine-tunes a complete RE10K-trained DepthSplat model on SceneSplat-Stereo.
Both are evaluated zero-shot on StereoNVS and the eight-scene ReplicaGS
benchmark; the released RE10K model is reported in Table~\ref{tab:main}.

Feature initialisation gives better ReplicaGS scores but worse StereoNVS
scores than the released RE10K model. RE10K fine-tuning improves all
three ReplicaGS metrics and StereoNVS PSNR, while StereoNVS SSIM/LPIPS
change from 0.7498/0.2024 to 0.7313/0.2057.
StereoGaussians exceeds both variants on all six test metrics,
supporting its effectiveness when the competing method also uses
SceneSplat-Stereo for view-synthesis training.

\clearpage
\section{Higher-resolution training and evaluation}
\label{app:resolution}

We additionally train the default architecture at $512\times512$ and
$1024\times1024$, initialising the trainable predictor independently
at each resolution. The pretrained stereo
encoder remains frozen. Both runs use the same optimiser recipe and
30\,000 training steps and validation-based model selection, and are
evaluated at their training resolution. Padding scales proportionally
from 16 pixels at $256^2$ to 32 at $512^2$ and 64 at $1024^2$.
Table~\ref{tab:resolution} also includes the $256^2$ / pad-64 variant
as a separate reference; its relative padding margin is larger.

\begin{table}[!ht]
\centering
\small
\caption{Higher-resolution reference results using the same model
architecture. Each predictor is trained from scratch for 30K steps
and evaluated at its training resolution, with a frozen pretrained
stereo encoder. Batch denotes effective batch size (GPUs $\times$
per-GPU batch). Padding scales with resolution except for the additional
$256^2$ / pad-64 reference.}
\label{tab:resolution}
\setlength{\tabcolsep}{3pt}
\resizebox{\linewidth}{!}{%
\begin{tabular}{lccccc|ccc}
\toprule
& & & \multicolumn{3}{c|}{StereoNVS} & \multicolumn{3}{c}{ReplicaGS} \\
Resolution & Pad & Batch & PSNR $\uparrow$ & SSIM $\uparrow$ & LPIPS $\downarrow$ & PSNR $\uparrow$ & SSIM $\uparrow$ & LPIPS $\downarrow$ \\
\midrule
$256^2$ (default) & 16 & $32\;(1\times32)$ & 21.261 & 0.7549 & 0.1653 & 30.266 & 0.9447 & 0.0818 \\
$256^2$ (larger FoV) & 64 & $32\;(1\times32)$ & 23.189 & 0.7689 & 0.1612 & 30.165 & 0.9443 & 0.0823 \\
\midrule
$512^2$ & 32 & $24\;(2\times12)$ & 20.414 & 0.7090 & 0.2274 & 29.676 & 0.9306 & 0.1061 \\
$1024^2$ & 64 & $32\;(8\times4)$ & 19.861 & 0.7183 & 0.2935 & 29.615 & 0.9324 & 0.1350 \\
\bottomrule
\end{tabular}%
}
\end{table}

The $512^2$ and $1024^2$ models attain 20.414 and 19.861\,dB on
StereoNVS, and 29.676 and 29.615\,dB on ReplicaGS, respectively.
These runs demonstrate support for higher-resolution stereo inputs
and target-view rendering with the same prediction architecture.
Each model is evaluated using targets at its own training resolution.
The $512^2$ run uses effective batch size 24;
the $256^2$ and $1024^2$ runs use 32. Table~\ref{tab:resolution}
reports all three quality metrics and the corresponding settings.

\section{Training-data preparation details}
\label{app:training-data}

This section details the IRS preparation used for the training-data
comparison in \S\ref{sec:training-data} and Table~\ref{tab:training-data}.

\paragraph{IRS camera reconstruction.}
IRS does not release camera extrinsics. We reconstruct each scene
variant independently with COLMAP from its left images, fixing the
intrinsics at $f_x=f_y=480$, $(c_x,c_y)=(480,270)$ for $960\times540$
images. We retain the largest reconstruction and recover metric scale
using the median ratio of ground-truth depth $z=48/d$ to reconstructed
depth at matched sparse points. Right-camera poses follow from the
rectified 0.1\,m baseline. We retain reconstructions with at least 150
registered frames, mean reprojection error at most 1.5 pixels, and
scale-ratio IQR at most 25\% of its median, yielding 63 scene variants.

\paragraph{IRS target selection and training.}
Targets are existing left-image frames from the original synthetic
trajectories. Reference frames are sampled at translation or rotation
increments of at least 5\,cm or $5^\circ$. Target candidates lie
3--40\,cm away with at most $30^\circ$ rotation and a projected-point
frustum coverage of at least 0.4. Coverage measures the fraction of
reference depth points projected in front of the target camera and
inside its image, without an occlusion test. Up to eight targets are
selected by proximity of this coverage to 0.8. The resulting pool has
47\,703 reference pairs and 303\,196 targets, split into 55 training
and eight validation scene variants. Training uses the 38\,286 training
pairs with at least four targets, sampling four per pair under the
default $256^2$, 30K-step, batch-32 recipe. Validation uses 40 pairs
and 160 targets for best-PSNR checkpoint selection.

SceneSplat-Stereo provides 803 training scenes and 323\,817 stereo
pairs. Its stronger ReplicaGS performance and lowest LPIPS on both
benchmarks (Table~\ref{tab:training-data}) demonstrate its effectiveness
as training data for cross-dataset generalisation and perceptual quality.

\clearpage
\section{Additional qualitative results}
\label{app:qualitative}

\subsection{Comparison with novel-view synthesis baselines}
\label{app:qualitative-comparison}

\begin{figure}[H]
\centering
\includegraphics[width=\linewidth]{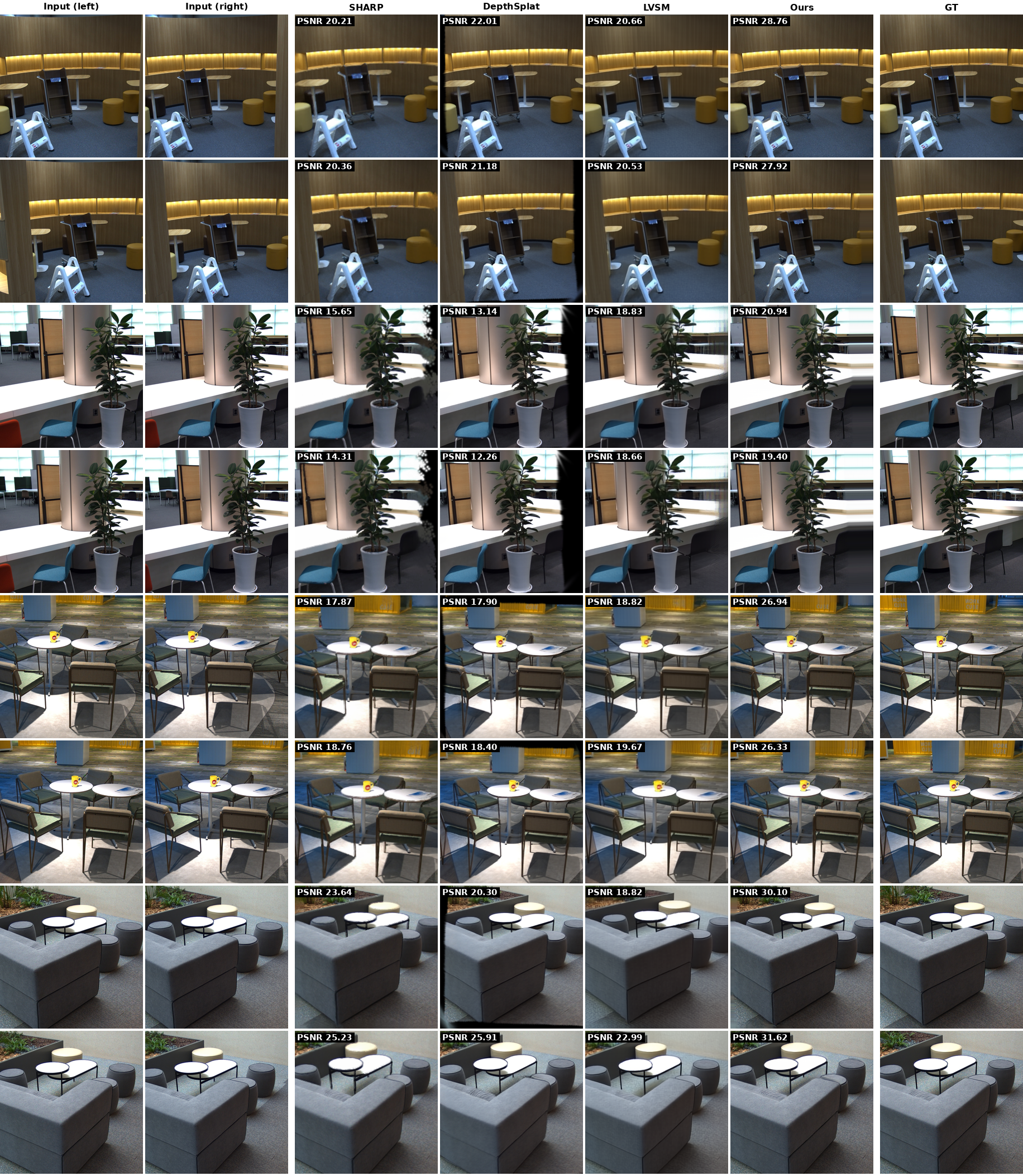}
\caption{\textbf{Additional baseline comparisons on StereoNVS (1/2).}
Columns show the left and right inputs, SHARP, DepthSplat, LVSM,
\titlePrefix{} (Ours), and ground truth (GT). Prediction panels report
PSNR in dB. In these lobby, plant, outdoor-seating, and sofa examples,
our model better preserves furniture contours and the separation
between foreground objects and their surroundings.}
\label{fig:app-comparison-stereonvs-1}
\end{figure}

\begin{figure}[H]
\centering
\includegraphics[width=\linewidth]{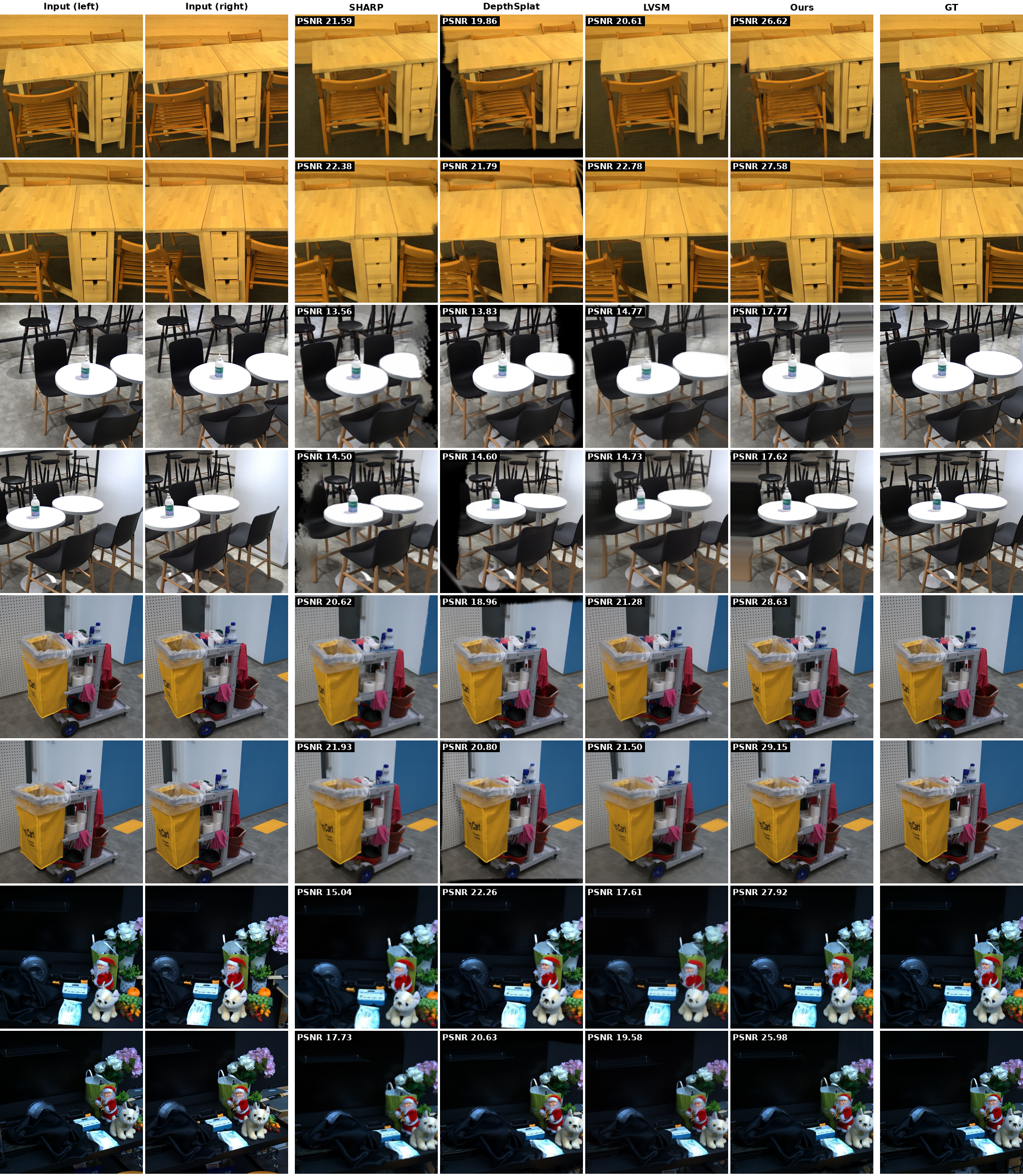}
\caption{\textbf{Additional baseline comparisons on StereoNVS (2/2).}
Column order and score annotations follow
Figure~\ref{fig:app-comparison-stereonvs-1}. These examples include
wooden desks, closely spaced chairs and tables, a cleaning cart, and
small decorative objects. Our predictions retain clearer object
boundaries and more of the small-scale structure in the cart and
decorations.}
\label{fig:app-comparison-stereonvs-2}
\end{figure}

\begin{figure}[H]
\centering
\includegraphics[width=\linewidth]{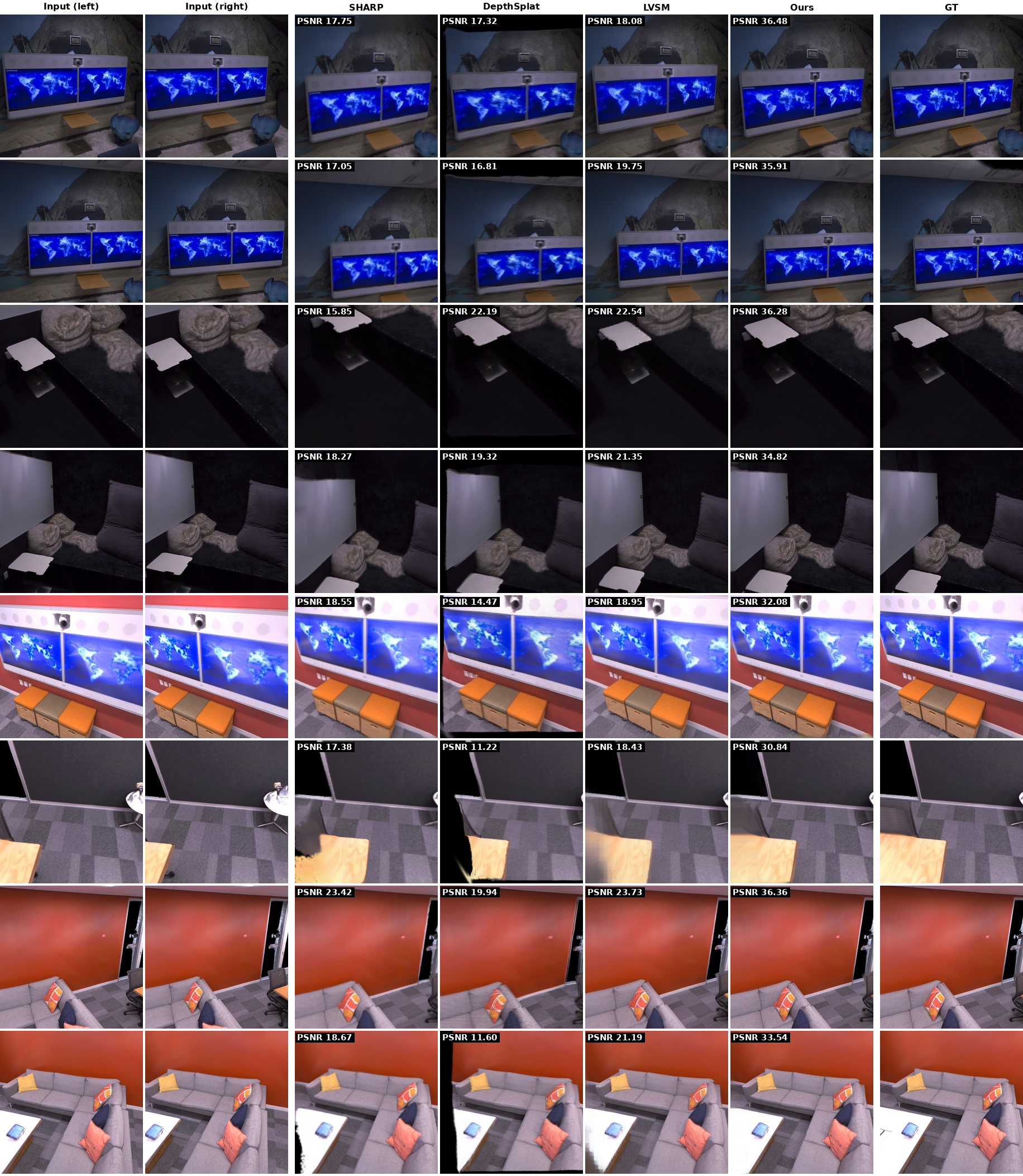}
\caption{\textbf{Additional baseline comparisons on ReplicaGS (1/2).}
Columns show the stereo inputs, SHARP, DepthSplat, LVSM,
\titlePrefix{} (Ours), and GT, with per-example PSNR in dB.
The display panels, cushions, floor patterns, and sofa boundaries
are more faithfully reproduced by our model in these examples.
Several baseline predictions exhibit displaced structures, blurred
textures, or missing content near image boundaries.}
\label{fig:app-comparison-replicags-1}
\end{figure}

\begin{figure}[H]
\centering
\includegraphics[width=\linewidth]{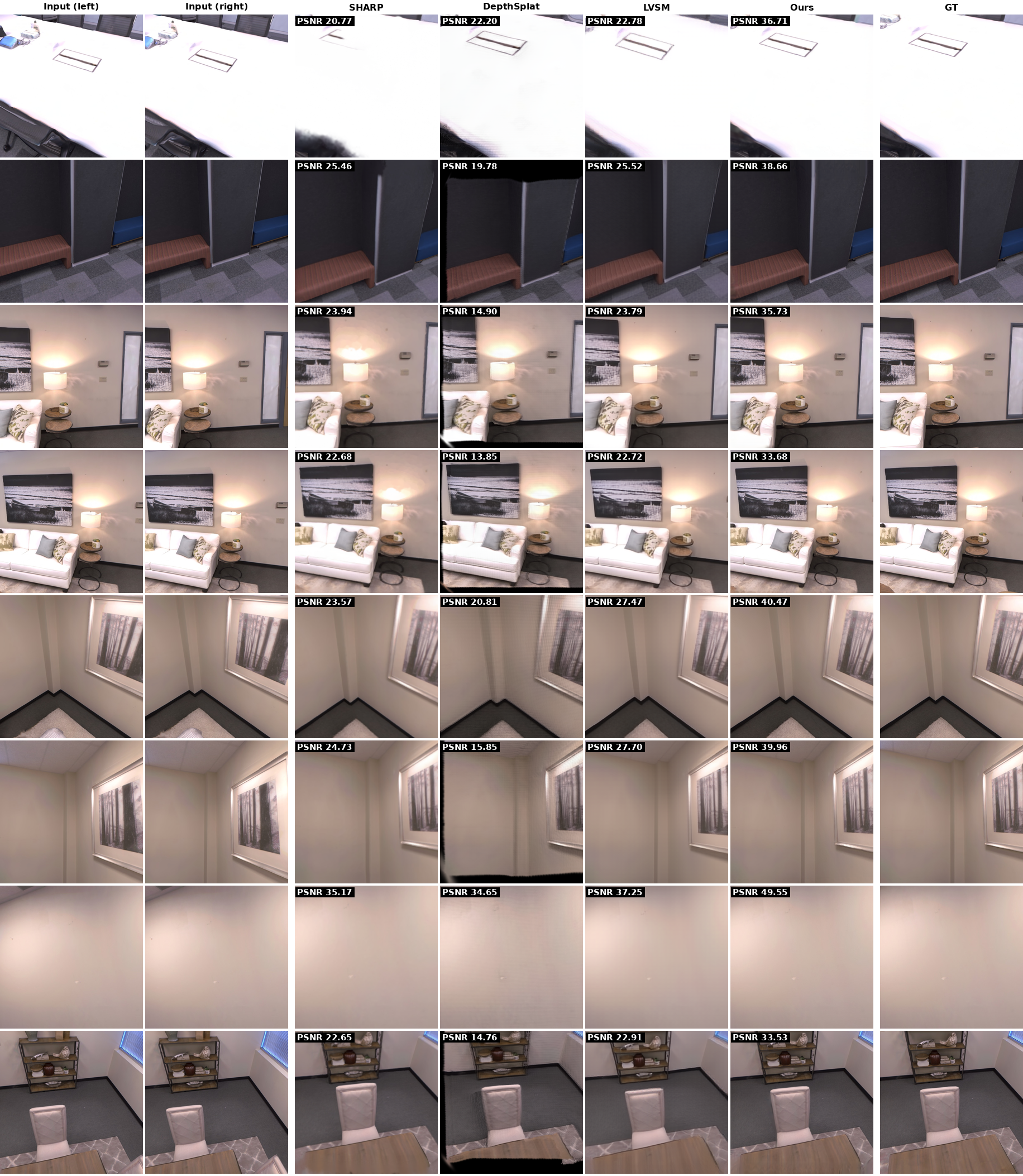}
\caption{\textbf{Additional baseline comparisons on ReplicaGS (2/2).}
Column order and score annotations follow
Figure~\ref{fig:app-comparison-replicags-1}. Examples cover tabletops,
wall partitions, a furnished living room, framed pictures, and a dining
area. Our model better maintains straight architectural edges and
furniture placement, while reducing the distortions and missing
border regions visible in several baseline renderings.}
\label{fig:app-comparison-replicags-2}
\end{figure}

Figures~\ref{fig:app-comparison-stereonvs-1}--\ref{fig:app-comparison-replicags-2}
extend the comparison in \S\ref{sec:main-comparison} with 16 examples
from each benchmark. Each row shows a stereo input pair and the
corresponding target-view predictions. The displayed PSNR values are
per-example scores; aggregate results are reported in Table~\ref{tab:main}.

\clearpage

\paragraph{Captured-image evaluation.}
The StereoNVS examples illustrate transfer to real captured imagery,
including clutter, thin structures, and changes in visibility.
The furniture and cleaning-cart views show improved boundary
definition and object structure, complementing the zero-shot
results in Table~\ref{tab:main}.

\clearpage

\clearpage

\paragraph{Teacher-rendered evaluation.}
ReplicaGS uses teacher-rendered targets from held-out scenes
(Appendix~\ref{app:eval-details}). The furniture contours, wall edges,
and floor patterns provide visual evidence of the rendering fidelity
reflected in Table~\ref{tab:main}.

\clearpage
\subsection{Qualitative ablations}
\label{app:qualitative-ablation}

\begin{figure}[H]
\centering
\includegraphics[width=\linewidth]{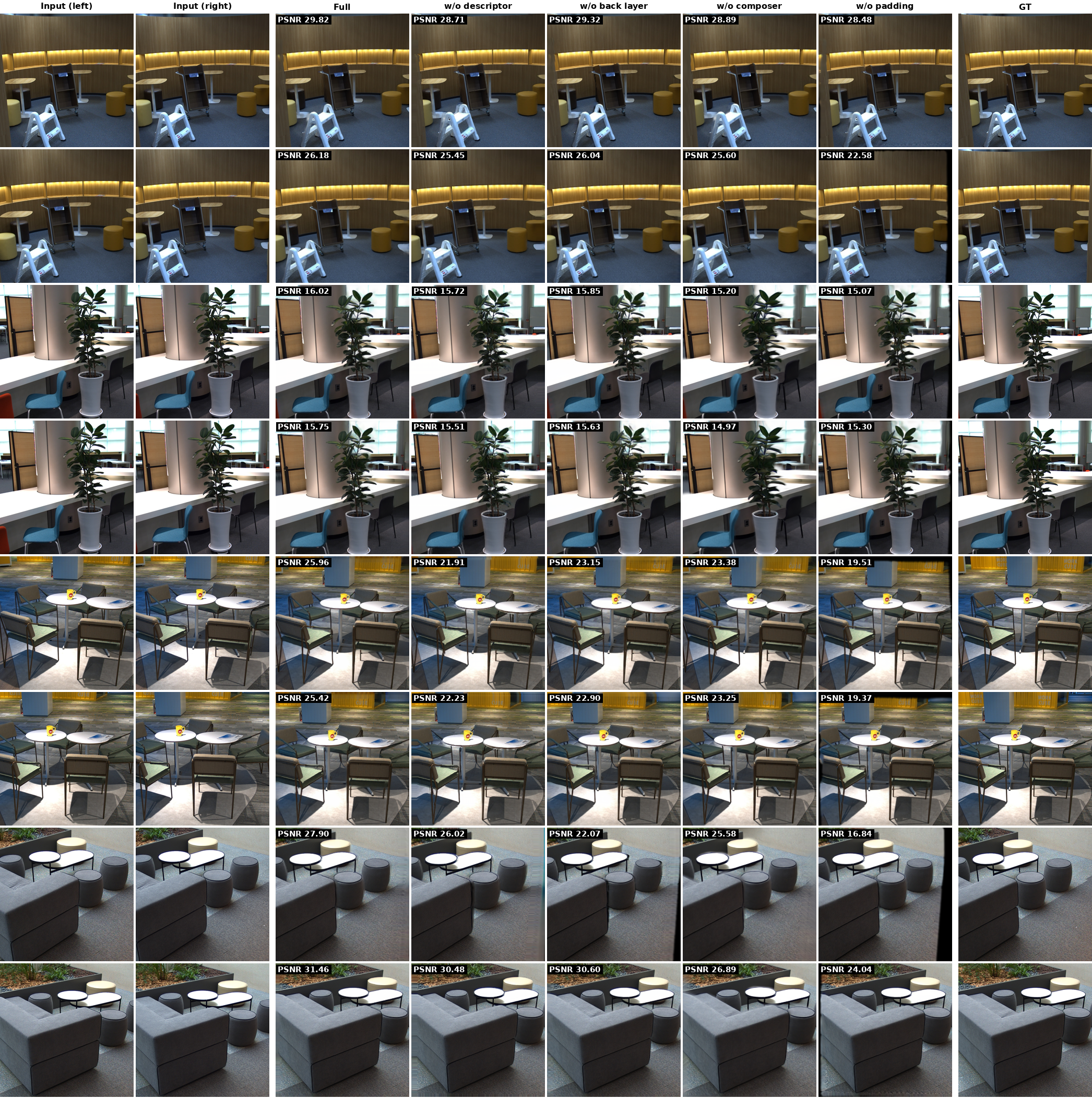}
\caption{\textbf{Qualitative ablations on StereoNVS (1/2).}
Columns show the stereo inputs, full model, four ablated variants,
and GT. Prediction panels report PSNR in dB. Removing the descriptor
or back layer degrades table and chair boundaries in the outdoor
examples. Removing padding leaves uncovered border regions,
particularly in the sofa views.}
\label{fig:app-ablation-stereonvs-1}
\end{figure}

\begin{figure}[H]
\centering
\includegraphics[width=\linewidth]{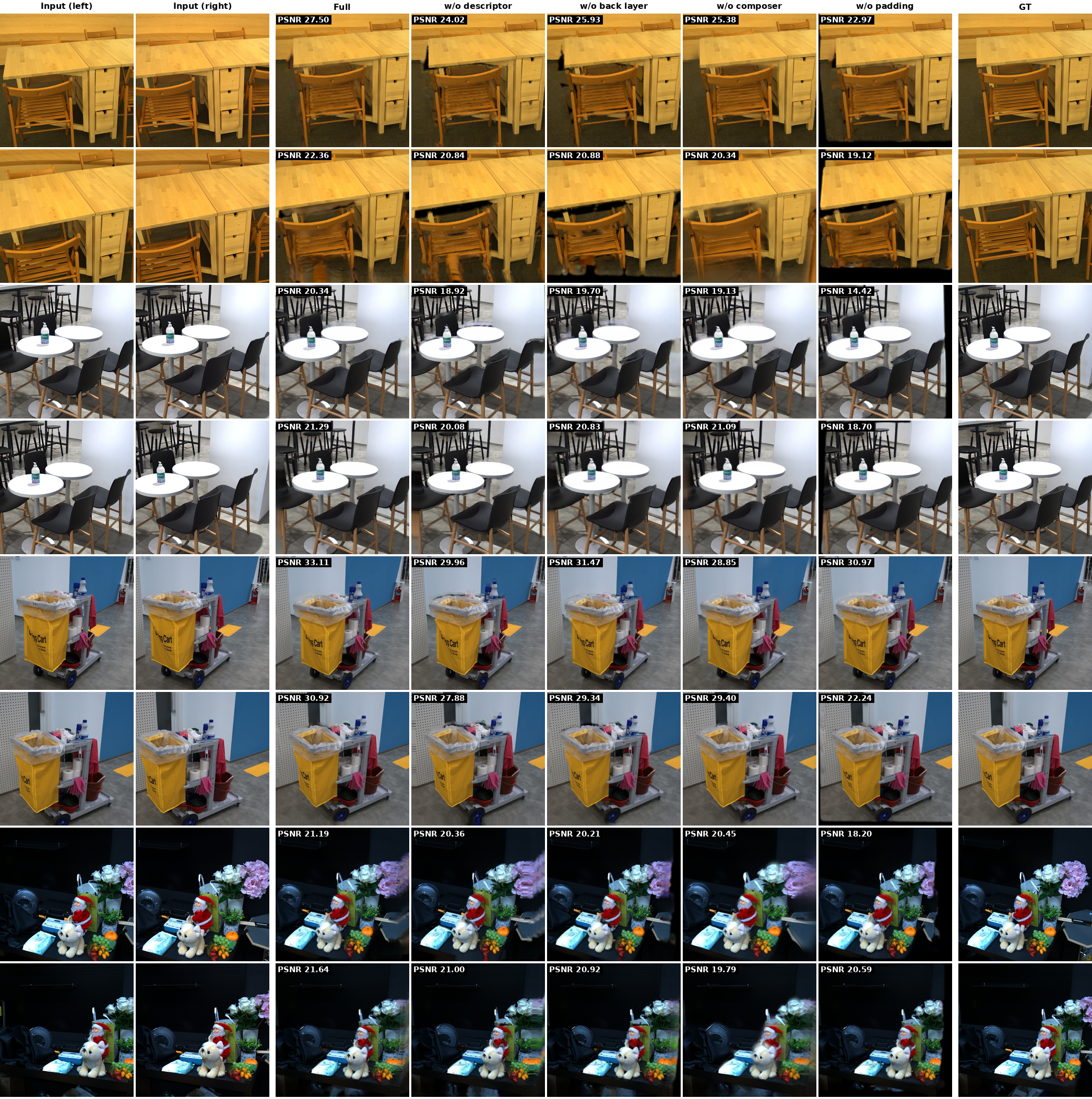}
\caption{\textbf{Qualitative ablations on StereoNVS (2/2).}
The layout follows Figure~\ref{fig:app-ablation-stereonvs-1}.
The desk and chair examples expose missing regions and distorted
boundaries in the ablated predictions. The cleaning cart and small
decorations show changes in fine structure and appearance. The full
model reduces these artifacts and better preserves the depicted
objects' structure.}
\label{fig:app-ablation-stereonvs-2}
\end{figure}

\begin{figure}[H]
\centering
\includegraphics[width=\linewidth]{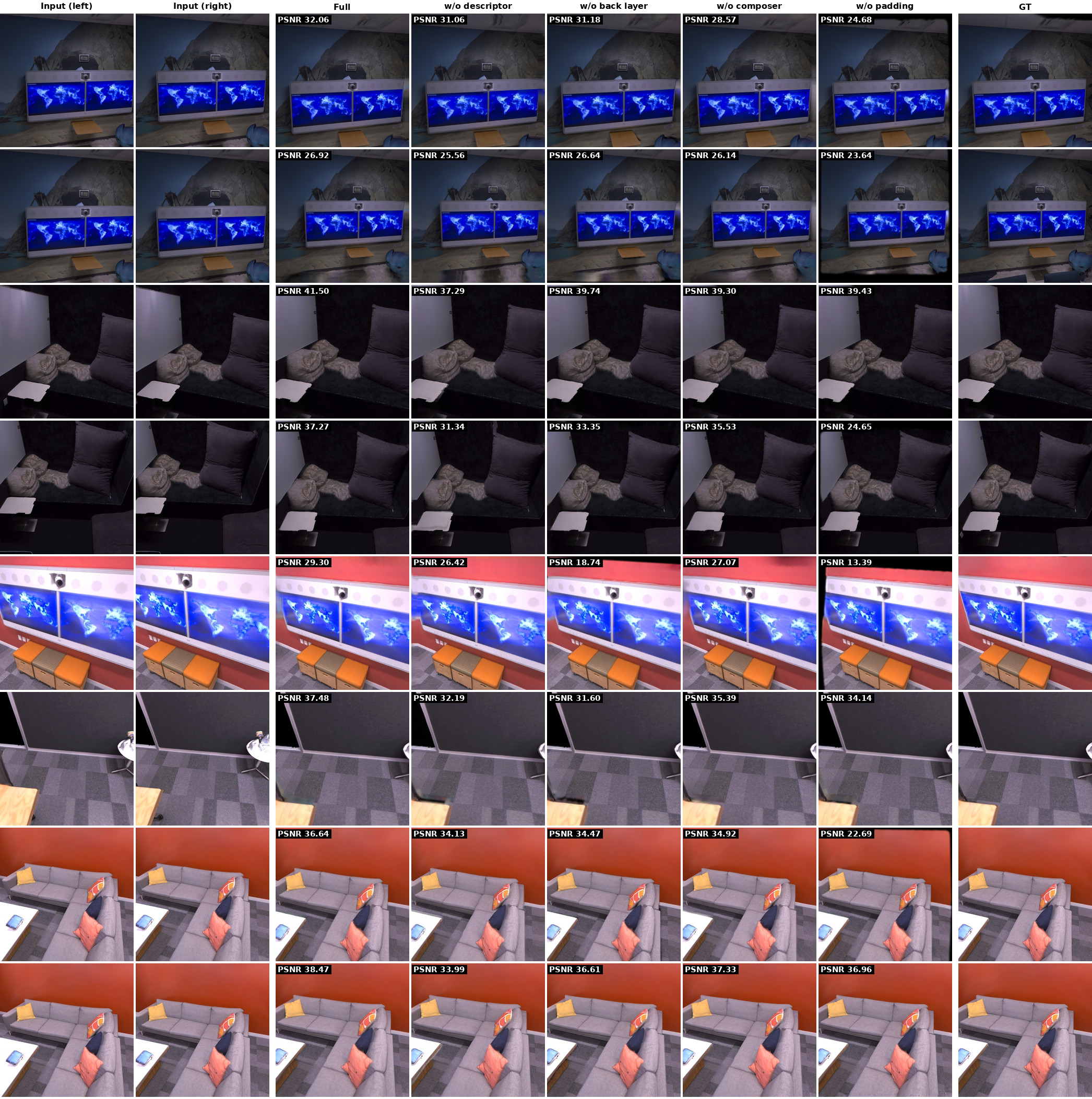}
\caption{\textbf{Qualitative ablations on ReplicaGS (1/2).}
Columns follow Figure~\ref{fig:app-ablation-stereonvs-1}, with
per-example PSNR in dB. Removing the descriptor reduces fidelity
around cushions and floor patterns. Removing the back layer can
introduce missing content, including the upper region of the bright
display scene. Without padding, black bands appear at newly exposed
image boundaries in several views.}
\label{fig:app-ablation-replicags-1}
\end{figure}

\begin{figure}[H]
\centering
\includegraphics[width=\linewidth]{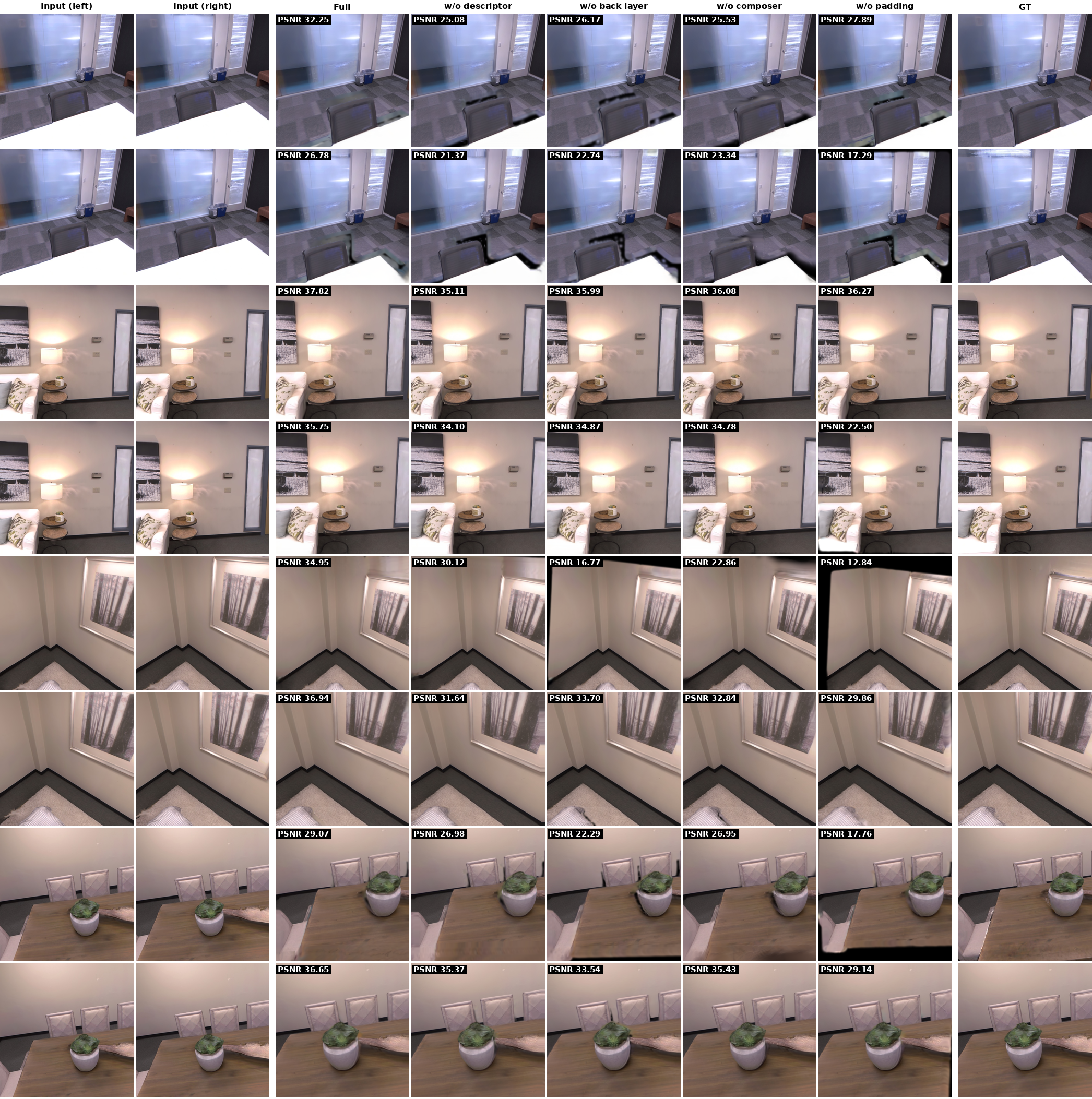}
\caption{\textbf{Qualitative ablations on ReplicaGS (2/2).}
The layout follows Figure~\ref{fig:app-ablation-stereonvs-1}.
The office-chair and dining-table views reveal gaps around foreground
objects when the descriptor or back layer is removed. The no-padding
variant additionally leaves large uncovered regions at the target
image borders. The full model improves coverage while retaining
clearer furniture and wall boundaries.}
\label{fig:app-ablation-replicags-2}
\end{figure}

Figures~\ref{fig:app-ablation-stereonvs-1}--\ref{fig:app-ablation-replicags-2}
complement Table~\ref{tab:ablations} and
Figure~\ref{fig:composer-diagnostic}. They compare the full model with
variants that remove the DPT descriptor, back Gaussian layer,
attribute composer, or canvas padding. Comparisons are made within
each row; the baseline and ablation grids do not always show the
same input or target views.

\paragraph{Feature conditioning.}
The descriptor-free variant retains RGB and stereo geometry but
loses the learned feature pathway into Gaussian prediction. The
tabletop and chair distortions illustrate that metric depth alone
does not determine the Gaussian attributes needed for faithful
target-view rendering.

\clearpage

\paragraph{Attribute composition.}
The composer improves local appearance and boundary quality around
the wooden furniture and small objects in these examples. These
visual improvements complement its lower LPIPS on both benchmarks
and higher StereoNVS SSIM (Table~\ref{tab:ablations}); the quantitative
comparison is discussed in \S\ref{sec:ablation-composer}.

\clearpage

\paragraph{Additional Gaussian capacity.}
The back layer adds capacity behind the stereo-anchored front layer.
Its removal can leave gaps when the target view reveals content
beyond foreground boundaries. The full model fills these regions
more completely in the displayed examples, consistent with the
ReplicaGS degradation measured for the single-layer variant in
Table~\ref{tab:ablations}.

\clearpage

\paragraph{Prediction-canvas coverage.}
Removing padding restricts Gaussian prediction to the original
reference-image canvas. The black bands along the office, wall,
and dining-table views demonstrate the resulting loss of coverage
under viewpoint changes. Extending the canvas reduces these gaps
and provides more complete target-view coverage, complementing the
PSNR, SSIM, and LPIPS improvements in Table~\ref{tab:ablations}.

\end{document}